\documentclass[10pt,twocolumn,letterpaper]{article}

\usepackage{cvpr}              
\usepackage{multirow}
\usepackage[table]{xcolor}
\usepackage{pifont}
\usepackage{array}
\definecolor{cvprblue}{rgb}{0.21,0.49,0.74}
\usepackage[pagebackref,breaklinks,colorlinks,allcolors=cvprblue]{hyperref}

\def\paperID{13529} 
\def\confName{CVPR}
\def\confYear{2026}

\title{GAIS: Frame-Level \underline{G}ated \underline{A}udio-Visual \underline{I}ntegration with \underline{S}emantic Variance-Scaled Perturbation for Text-Video Retrieval}

\author{Bowen Yang, Yun Cao, Chen He, Xiaosu, Su\\
Institue of Information Engineering, Chinese Academy of Sciences, Beijing, China\\
School of Cyber Security, University of Chinese Academy of Sciences, Beijing, China\\
{\tt\small \{yangbowen, caoyun, hechen, suxiaosu\}.iie.ac.cn}
}

\begin{document}
\maketitle


\begin{abstract}
    Text-to-video retrieval requires precise alignment between language and temporally rich audio-video signals. However, existing methods often emphasize visual cues while underutilizing audio semantics or relying on coarse fusion strategies, resulting in suboptimal multimodal representations. We introduce \textbf{GAIS}, a retrieval framework that strengthens multimodal alignment from both representation and regularization perspectives. First, a Frame-level Gated Fusion (FGF) module adaptively integrates audio-visual features under textual guidance, enabling fine-grained temporal selection of informative frames. Second, a Semantic Variance-Scaled Perturbation (SVSP) mechanism regularizes the text embedding space by controlling perturbation magnitude in a semantics-aware manner. These two modules are complementary: FGF minimizes modality gaps through selective fusion, while SVSP improves embedding stability and discrimination. Extensive experiments on MSR-VTT, DiDeMo, LSMDC, and VATEX demonstrate that GAIS consistently outperforms strong baselines across multiple retrieval metrics while maintaining notable computational efficiency. 
\end{abstract}

\begin{figure}[ht]
    \centering
    \includegraphics[width=1\linewidth]{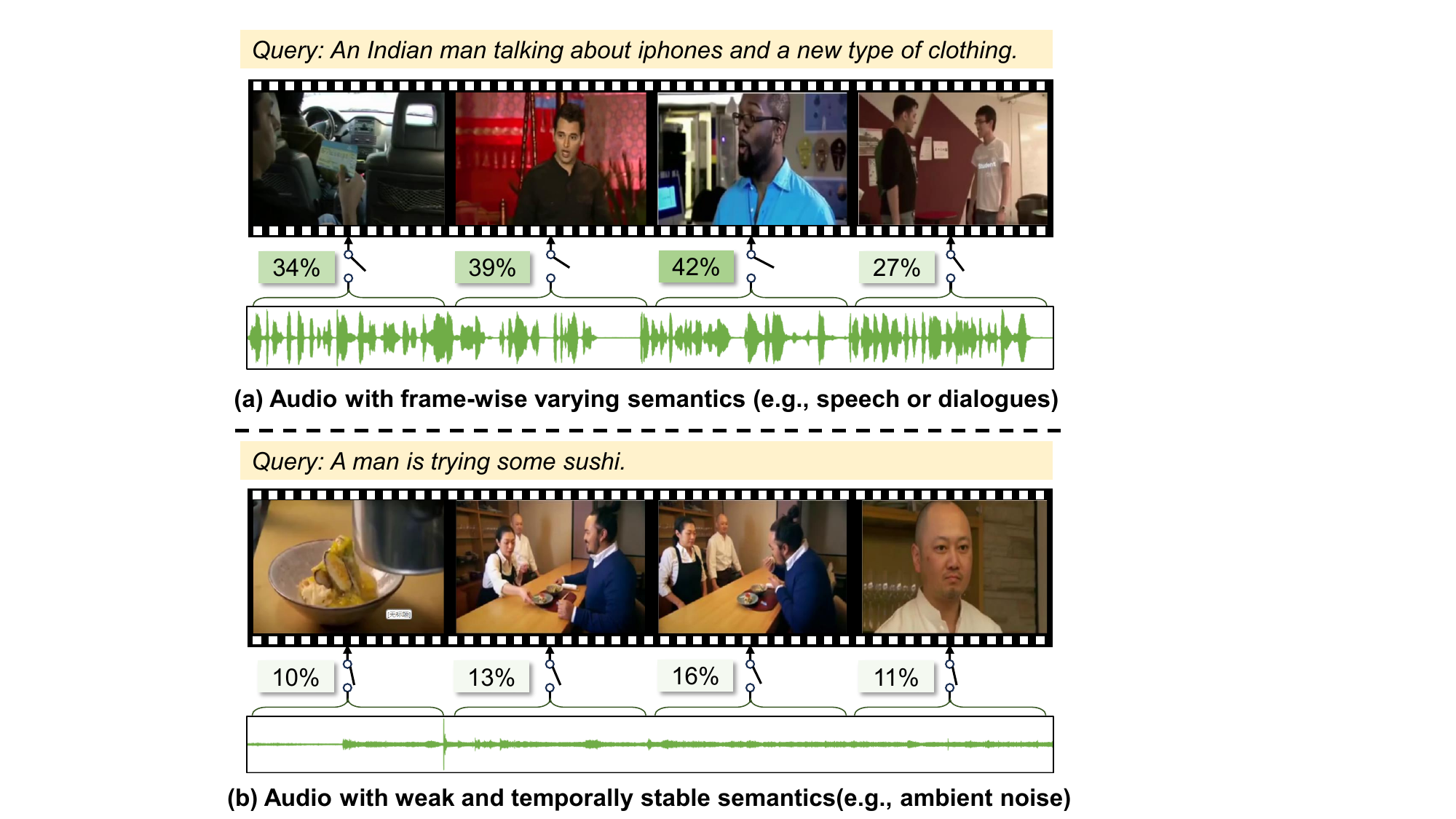}
    \caption{Illustration of Frame-level Gated Fusion. (a) When audio contains salient semantic cues, the gate assigns higher weights to audio across relevant frames. (b) When the audio is dominated by background noise, the gate suppresses audio contributions, preventing irrelevant signals from affecting retrieval.}
    \label{fig:retrieval_result}
\end{figure}

\section{Introduction}

Text-to-video retrieval (T2VR) aims to match natural language descriptions with temporally structured video content. While recent dual-encoder architectures \cite{bain_frozen_2021, DBLP:conf/cvpr/querybank, DBLP:journals/corr/dual_softmax, luo_clip4clip_2021} have achieved impressive progress, they still face two fundamental challenges: (1) the inability to capture fine-grained temporal relevance across multimodal signals, and (2) the lack of robust semantic regularization for noisy cross-modal correspondence. 

Videos are inherently redundant, and only a small subset of frames aligns semantically with the textual query \cite{DBLP:conf/sigir/centerclip, xue_clip-vip_2023, wu_cap4video_2023, wang_text_2024}. Moreover, audio cues often encode complementary semantics such as speech, ambient context, or emotional tone, which purely visual models ignore. Existing approaches either omit audio entirely or fuse it in a static or coarse-grained manner \cite{DBLP:conf/sigir/centerclip, xue_clip-vip_2023, wu_cap4video_2023, wang_text_2024}, producing over-smoothed representations that blur discriminative details and degrade retrieval accuracy.

To overcome these limitations, we propose GAIS,  a frame-level gated audio-visual integration framework with semantic variance-scaled regularization.  As shown in \cref{fig:retrieval_result}, audio cues may be highly informative or entirely uninformative depending on the temporal context. This observation motivates our frame-level gated fusion module, which selectively enhances semantically aligned audio segments while suppressing irrelevant or redundant ones. By focusing on informative moments, GAIS enables fine-grained temporal alignment and yields stronger cross-modal correspondence.

Another key challenge in T2VR lies in representation robustness. Prior work \cite{wang_text_2024} has attempted to introduce stochastic perturbation to text embeddings to improve generalization. However, such noise injection typically has uncontrolled magnitude and may distort semantic meaning. Moreover, these methods often require multiple random samples at inference time to achieve stable similarity estimation, leading to inconsistency between training and inference as well as increased computational overhead.

To overcome these issues, we propose a Semantic Variance-Scaled Perturbation mechanism that regulates perturbation magnitude according to the cross-modal variance implied by video features. During training, the model maintains mild stochasticity for robustness while preventing excessive deviation from the semantic anchor. During inference, the perturbation becomes deterministic, enabling stable single-pass retrieval without sampling. This design allows the model to benefit from perturbation-driven regularization while preserving semantic consistency and efficient inference.

In summary, our contributions are as follows:

\begin{itemize}
    \item We introduce a Frame-level Gated Fusion module that adaptively integrates audio and visual features under textual guidance, enabling fine-grained temporal alignment.
    \item We propose a Semantic Variance-Scaled Perturbation mechanism that learns to control perturbation magnitude while preserving semantic consistency, providing robust training and deterministic single-pass inference.
    \item Extensive experiments on MSR-VTT, DiDeMo, LSMDC, and VATEX demonstrate that the proposed framework consistently improves retrieval performance and maintains favorable efficiency.
\end{itemize}

\section{Related Work}

\subsection{Text-Video Retrieval}
Early approaches to T2VR primarily focused on visual signals and relied on multi-level semantic alignment to bridge the cross-modal gap. Classical methods such as hierarchical matching frameworks \cite{DBLP:conf/cvpr/ChenZJW20,DBLP:conf/mm/WuHTLL21} and multi-stream frameworks like MTVR \cite{DBLP:conf/eccv/Gabeur0AS20} and T2VLAD \cite{DBLP:conf/cvpr/WangZ021}, which capture actions, objects, and scenes through hand-crafted combinations of local and global features. While these designs established the foundation of T2VR, they were limited by their weak temporal modeling and the lack of end-to-end optimization.

With the rise of large-scale vision-language pretraining, the field shifted toward jointly learned embeddings. Models such as ClipBERT \cite{DBLP:conf/cvpr/LeiLZGBB021} and Frozen \cite{bain_frozen_2021} pioneered joint pretraining for video-text tasks, followed by CLIP4Clip \cite{luo_clip4clip_2021}, which directly transfers CLIP \cite{DBLP:conf/icml/RadfordKHRGASAM21} embeddings to retrieval. Later works like TS2-Net \cite{liu_ts2-net_2022} and DRL \cite{DBLP:journals/corr/drl-wang} improved temporal reasoning through token shift-selection and disentangled hierarchical modeling. However, most of these visual-centric methods remain audio-agnostic and treat the video as a purely visual signal. 

\begin{figure*}[!ht]
    \centering
    \includegraphics[width=1\linewidth]{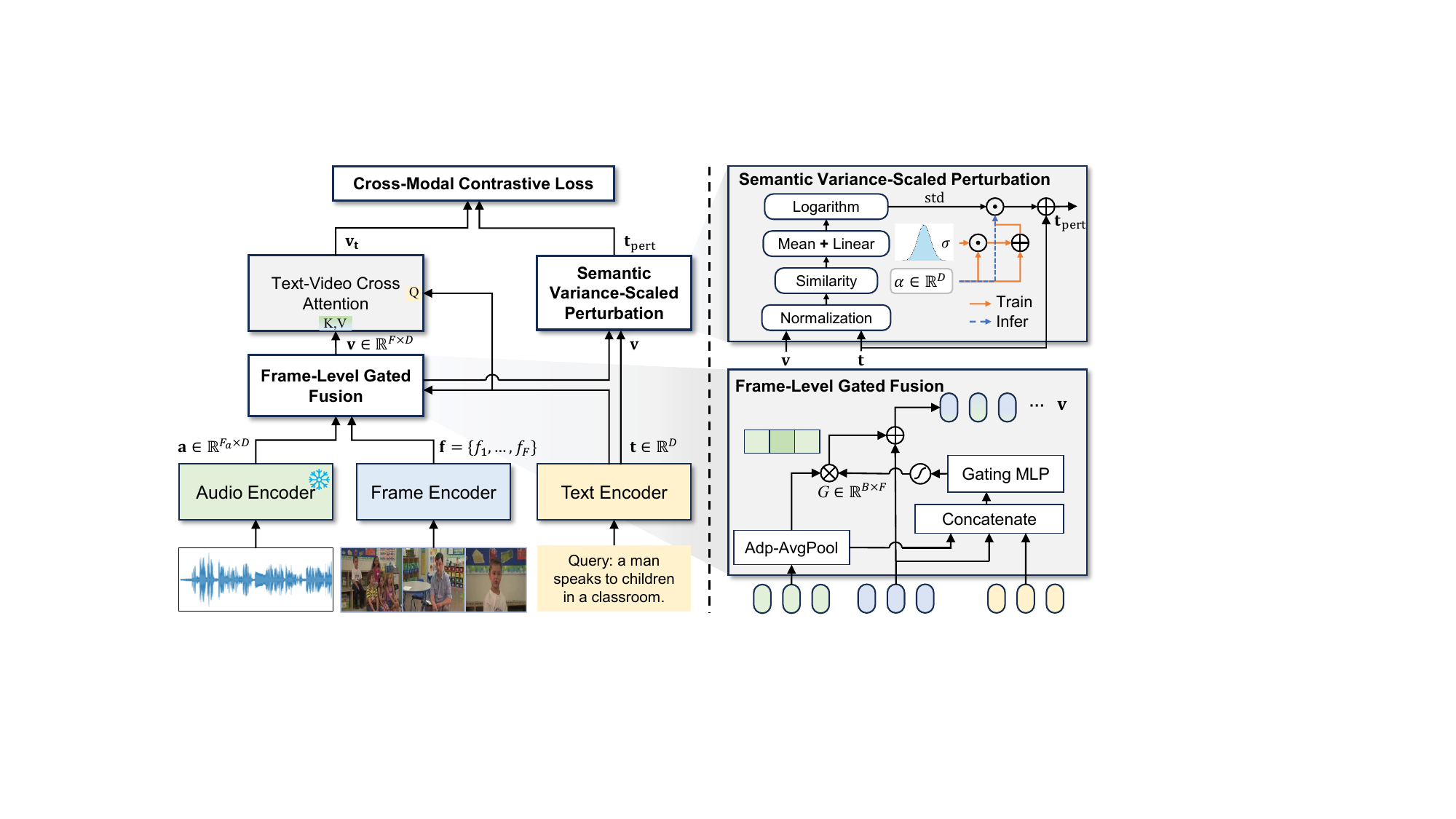}
    \caption{Overview of GAIS. Given video frames, audio, and a text query, Frame-level Gated Fusion (FGF) adaptively integrates audio-visual features conditioned on text. The fused features are enhanced via text-video cross-attention and fed into the Semantic Variance-Scaled Perturbation (SVSP) module. Training uses stochastic perturbation for regularization, while inference employs a single deterministic pass for efficiency. $G$ represents the frame-level gated feature matrix (batch$\times$frames).}
    \label{fig:main_method}
\end{figure*}

Recently, robustness-focused methods emerged. T-MASS \cite{wang_text_2024} enhances discrimination by introducing stochastic perturbations to text embeddings, requiring multiple inference passes. InternVid \cite{DBLP:conf/iclr/InternVid} scales pretraining to massive datasets and introduces ViCLIP with spatiotemporal attention. In contrast, GAIS redefines both \textit{representation} and \textit{regularization} aspects of T2VR. 

\subsection{Audio in Multimodal Learning}

Incorporating audio as a complementary modality has gained traction in multimodal learning. Early methods \cite{DBLP:journals/corr/abs-1804-02516, DBLP:conf/nips/AkbariYQCCCG21, DBLP:conf/nips/AlayracRSARFSDZ20} aligned visual, auditory, and textual using self-supervised training but were constrained by weak audio encoders and limited semantic richness.

More recent audio-aware retrieval models adopt cross-modal fusion mechanisms. ECLIPSE \cite{eccv_EclipSE} introduced symmetric cross-attention between audio and video streams, while TEFAL \cite{audio_enhanced} adopted text-conditioned cross-attention for multimodal fusion. AVIGATE \cite{DBLP:conf/cvpr/JeongPKK25} further introduced a multi-layer gated fusion strategy, hierarchically combining audio and visual features for richer interactions. VALOR \cite{DBLP:journals/pami/valor},  and VAST \cite{DBLP:conf/nips/VAST}, explore large-scale audio-visual-text pretraining. However, these methods typically employ sample-level fusion or token-level attention, either missing fine-grained temporal dynamics or incurring heavy computation.  

Existing research reveals two gaps: (i) fusion granularity—most audio-aware methods operate at sample-level or token-level extremes, either ignoring dynamic temporal variation or introducing high computational overhead; and (ii) text regularization—robustness enhancements via stochastic perturbation require costly multi-sampling and lack structural guidance from cross-modal signals. GAIS addresses these gaps through a frame-level text-guided gating mechanism for audio-visual fusion and a deterministic directional perturbation for robust text embeddings, jointly enabling fine-grained alignment with low inference cost.

\section{Methodology}

We propose GAIS, a unified framework that enhances multimodal alignment from both representation and regularization perspectives. Given a text query $\mathbf{t}$ and video containing both visual frames and audio signals, GAIS learns to embed them into a shared semantic space while addressing two challenges: (1) identifying when and where audio cues contribute to visual semantics, and (2) ensuring semantic stability of text embeddings under multimodal variance. 

\textbf{Overview.} As illustrated in \cref{fig:main_method}, GAIS follows a four-stage process: 
\begin{enumerate}
    \item The input video is decomposed into visual frames $\mathbf{f}$ and synchronized audio segments $\mathbf{a}$. Visual features are extracted using a CLIP-based \cite{DBLP:conf/icml/RadfordKHRGASAM21} image encoder, while audio waveforms are transformed into log-Mel spectrograms and encoded via Whisper \cite{DBLP:conf/icml/RadfordKXBMS23}. This produces temporally aligned frame-audio pairs. 
    \item Then, these paired features are fused under textual guidance through a learnable gating mechanism. For each frame, FGF determines the optimal balance between visual and audio cues, producing refined multimodal features $\mathbf{v} $ that emphasize semantically relevant frames and suppress redundant noise. 
    \item Third, SVSP refines the text embedding $\mathbf{t}$ using a cross-modal variance signal estimated from the video representation $\mathbf{v}$, enabling the perturbation magnitude to adapt to the uncertainty reflected in the text–video similarity.
    \item Fourth, like X-Pool \cite{DBLP:conf/cvpr/X-PoolGort}, a cross-attention pooling layer is introduced to achieve cross-modal refinement. Specifically, the text embedding $\mathbf{t}$ acts as query attending over the temporal multimodal features, forming an interaction-enhanced video embedding:
    \begin{equation}
        \mathbf{v_t}=\text{softmax}(\frac{\mathbf{t}W_q(\mathbf{v}W_k)^T}{\sqrt{d}})\mathbf{v}W_v,
    \end{equation}
    where $W_q$,$W_k$, $W_v$ are learnable projections. This pooling operation enables cross-modal refinement, allowing the text query to adaptively aggregate video evidence most relevant to its semantics, effectively bridging the local–global gap before contrastive optimization. 
\end{enumerate}
Finally, both text embedding $\mathbf{t}_\text{pert}$ and video embedding $\mathbf{v_t}$are projected into a shared retrieval space and jointly via a contrastive objective. This pipeline allows GAIS to capture fine-grained temporal dependencies while maintaining semantic consistency across modalities, leading to more accurate and interpretable retrieval.

\subsection{Frame-level Gated Audio-Visual Fusion}\label{sec:FGF}

Most prior works perform audio-visual fusion at the clip or token level, which fails to account for temporal relevance. As a result, background sounds (\textit{e.g.}, ambient noise, background music) may dominate informative cues. To mitigate this, we design a text-guided gating mechanism that adaptively determines the contribution of each modality at each frame. We denote the $i$-th sample's frame-level gated weight as $\text{G}_i=\{g^i_1, ...,g^i_F\}$.
Before fusion, we first align audio to the visual frame rate via adaptive avergae pooling ($F_a\rightarrow F$), ensuring that each frame has a corresponding audio segment.

Given frame features $f_i\in \mathbb{R}^D$ and temporally aligned audio features $a_i\in \mathbb{R}^D$. For each timestamp $t$, we compute a relevance gate:
\begin{equation}
    \begin{aligned}
        g_t = \sigma(W_g [f_t; a_t; \mathbf{t}]),
    \end{aligned}
\end{equation}

where $\sigma$ is a sigmoid function and $W_g$ are learnable projections. The gate $g_i\in [0,1]$ assigns higher weights to frames whose audio-visual cues correlate with the textual description, and suppresses irrelevant or background segments.

The gated fused representation is then obtained as:
\begin{equation}
    \begin{aligned}
        v_t = g_t \cdot a_t + (1-g_t)\cdot f_t,
    \end{aligned}
\end{equation}
enabling frame-specific weighting between modalities.

Unlike \cite{DBLP:conf/cvpr/JeongPKK25}, which applies static or layer-wise gating across all frames, FGF operates dynamically at the frame level under textual supervision.
This allows the model to amplify discriminative audio cues (\textit{e.g.}, speech, impact sounds) only when they align with the linguistic intent.

\subsection{Semantic Variance-Scaled Perturbation}\label{SVSP}

While FGF determines which frames contribute meaningful evidence, alignment strength across semantic dimensions can still vary depending on how specific or ambiguous the text description is. Some textual attributes (\textit{e.g.}, object identity) are visually grounded and stable, while others (\textit{e.g.}, action style or scene context) may align loosely with multiple plausible interpretations. To model this variation, we introduce SVSP, which adjusts the text embedding according to cross-modal semantic variance with dimension-wise learnable magnitude control.

To estimate the semantic variance used in SVSP, we compute a text-conditioned similarity signal between normalized text and video embeddings. Each text embedding is replicated across frames, frame-level similarities are averaged, and the result is transformed into a variance vector by a linear layer followed by a logarithm:
\begin{equation}
    \text{std} = \text{log}(\text{Linear}(\frac{1}{F}\sum_{f=1}^F s^{(f)}_{ij})),
\end{equation}
where $s^{(f)}_{ij}$ means the similarity between the $i$-th video frame and $j$-th text embedding. This produces a $D$-dimensional variance vector for each text–video pair, reflecting cross-modal uncertainty and controlling the perturbation magnitude in SVSP.

Let $\mathbf{t}\in\mathbb{R}^D$ denote the normalized text embedding. If we directly inject random perturbations, the embedding may deviate in arbitrary directions and with uncontrolled magnitude, which can distort semantics and lead to unstable similarity scores. Instead, we perturb $\mathbf{t}$ only along the semantic variance dimensions, and regulate how far the embedding is allowed to move.

During training, we introduce lightweight stochasticity while controlling the perturbation magnitude through a learnable coefficient $\alpha \in \mathbb{R}^D$  :
\begin{equation}
    \begin{aligned}
        \mathbf{t}_\text{pert}=\mathbf{t} + (\alpha \odot \sigma + (1-\alpha)) \odot \text{std}, \quad \sigma \sim \mathcal{N}(0, I),
    \end{aligned}
\end{equation}
\noindent where $\odot $ denotes element-wise multiplication. The term $\alpha \odot \sigma + (1-\alpha)$ balances robustness and stability: the stochastic component promotes generalization, while the residual deterministic term prevents excessive deviation from the semantic center. This maintains representation flexibility without causing semantic drift.

During inference, the perturbation becomes deterministic:
\begin{equation}
    \begin{aligned}
        \mathbf{t}_\text{pert} = \mathbf{t} + \alpha \odot \text{std}.
    \end{aligned}
\end{equation}

This enables a single forward pass and eliminates the multi-sample inference required by earlier stochastic perturbation methods, yielding consistent similarity estimates with improved retrieval efficiency.

Theoretically, SVSP reduces similarity variance by adjusting only dimensions associated with semantic uncertainty, avoiding uncontrolled drift. Geometrically, this corresponds to an ellipsoidal local neighborhood aligned with cross-modal variance, rather than an isotropic noise cloud.

 

Together, FGF and SVSP address two complementary aspects of cross-modal alignment:
FGF provides selective grounding by focusing on the most relevant audio–visual frames,
while SVSP provides confidence-adjusted matching by regulating how strongly the text embedding should align with the visual signal.
This combination yields robust and discriminative retrieval representations. We provide additional visualizations of gating behavior (\cref{fig:gate_vis}) and perturbation geometry (\cref{fig:SVSP_vis}) to illustrate the interpretability and stability of our modules.

\subsection{Training Objective}

We optimize the model using a contrastive retrieval loss applied to the perturbed text representation. Given the SVSP-augmented text embedding $\mathbf{t_pert}$ and pooled video embedding $\mathbf{v}$, the similarity matrix is computed as: 
\begin{equation}
\begin{aligned}
\mathbf{s}_{ij}=<\mathbf{t}^{(i)}_\text{pert}, \mathbf{v}^{(j)}>.
\end{aligned} 
\end{equation}

A temperature-scaled contrastive loss is applied symmetrically to encourage consistent video-to-text and text-to-video alignment. 

Following prior work \cite{wang_text_2024}, we additionally adopt a lightweight support-based refinement term to stabilize margin learning. The idea is to construct a controlled hard positive by shifting the text embedding toward its paired video embedding along the semantic alignment direction:

\begin{equation}
\begin{aligned}
\mathbf{t}_\text{sup}=\mathbf{t}+\frac{\mathbf{v}-\mathbf{t}}{||\mathbf{v}-\mathbf{t}||}\odot \text{exp}(\text{std}).
\end{aligned} 
\end{equation}

This refinement does not introduce new architectural components and is not a main contribution of our framework; rather, it acts as a complementary stabilizer to SVSP. While SVSP regularizes the text representation according to semantic variance, the support-based sample provides a slightly more challenging positive that encourages consistent alignment when the textual cues are weak or ambiguous. A contrastive loss computed on $\mathbf{t}_\text{sup}$ improves boundary stability with negligible computational overhead.

We adopt the standard bidirectional contrastive loss used in text-video retrieval:
\begin{equation}
\mathcal{L}_\textbf{t}=\mathcal{L}_{t\rightarrow v}+\mathcal{L}_{v\rightarrow t}
\end{equation}
where each term is an InfoNCE objective computed over the batch.
The final objective is:
\begin{equation}
    \begin{aligned}
    \mathcal{L}_{\text{total}} = \mathcal{L}_{\textbf{t}} + \lambda \mathcal{L}_{\textbf{t}_\text{sup}}.
    \end{aligned}
\end{equation}
where $\lambda$ balances robustness and boundary shaping.

This boundary refinement mechanism is a standard strategy used in cross-modal matching, and in our framework, it complements SVSP by ensuring that the learned embedding neighborhood remains compact and discriminative. 


\section{Experiment}

\textbf{Datasets and Metrics.} We adopt four benchmark datasets for the evaluation, including (1) \textbf{MSR-VTT} \cite{DBLP:conf/cvpr/XuMYR16} is the most common dataset for text-to-video retrieval and the videos come with an audio track, consisting of 10,000 web video clips, each associated with 20 textual descriptions, we train GAIS on 9,000 videos and evaluate it on 1,000 selected pairs. (2)\textbf{LSMDC} \cite{DBLP:conf/cvpr/RohrbachRTS15} contains 118,081 video clips collected from 202 movies, with each clip paired with a textual description. Video lengths range from 2 to 30 seconds, and the dataset is split into 101,079 training, 7,408 validation, and 1,000 testing samples, following the setting of \cite{DBLP:conf/cvpr/X-PoolGort}. (3)\textbf{DiDeMo} \cite{DBLP:conf/iccv/HendricksWSSDR17} consists of 10,642 video clips and 40,543 textual descriptions; (4)\textbf{VATEX} \cite{DBLP:conf/iccv/WangWCLWW19} contains 34,991 video clips with multiple textual descriptions for each video.

We report Recall@K (R@1/5/10), Median Rank (MdR) and Mean Rank (MnR). Higher R@K, lower MdR and MnR indicate better performance. 

\textbf{Implementation Details.} For video frames and texts, we use CLIP \cite{DBLP:conf/icml/RadfordKHRGASAM21}’s visual and textual encoders (both ViTB/32 and ViT-B/16) to capture the respective modalities. For audio, we leverage open-source automatic speech recognition models \cite{DBLP:conf/icml/RadfordKXBMS23, DBLP:conf/nips/BaevskiZMA20} to encode raw audio signals into fixed-dimensional embeddings, which are temporally down-sampled via average pooling to match the 12 uniformly sampled video frames \cite{luo_clip4clip_2021}. For videos lacking audio, zero vectors are inserted to preserve modality alignment. All features are projected to a 512-dimensional space and fine-tuned with batch size 32, weight decay 0.2, and 5 epochs. Training is conducted on 1 to 4 NVIDIA L40 GPUs. Additional implementation details are provided in the Appendix~\ref{sec:impl_details}.

\begin{table*}[htbp]
\centering
\fontsize{9}{10}\selectfont

    \begin{tabular}{l|c|lllll|lllll}
        \toprule
        \multirow{ 2}{*}{Method}  &\multirow{ 2}{*}{Modality}& \multicolumn{5}{c|}{MSR-VTT Retrieval} &  \multicolumn{5}{c}{DiDeMo Retrieval} \\
        
         && R@1$\uparrow$ & R@5$\uparrow$ & R@10$\uparrow$ & MdR$\downarrow$ & MnR$\downarrow$  & R@1$\uparrow$ & R@5$\uparrow$ & R@10$\uparrow$ & MdR$\downarrow$ & MnR$\downarrow$ \\
        \midrule
        \textit{\textcolor{OliveGreen}{ViT-B/32}}  &&\multicolumn{5}{l|}{} &\multicolumn{5}{l}{} \\
        CLIP4Clip \cite{luo_clip4clip_2021} &V+T&43.1    &70.4    &80.8  &2.0 &15.3  &43.4    &73.2    &80.6  &2.0 &21.6  \\
        ECLIPSE \cite{eccv_EclipSE}       &A+V+T&44.2    &71.3    &81.6  &2.0 &15.0  &44.2    &-       &-     &-   &-  \\
        BridgeFormer \cite{DBLP:conf/cvpr/bridgeformerGeGLLSQL22}   &V+T&44.9    &71.9    &80.3  &2.0 &15.3  &37.0    &62.2    &73.9  &3.0 &-  \\
        X-CLIP \cite{DBLP:conf/mm/X-CLIPMaXSYZJ22}         &V+T&46.1    &73.0    &83.1  &2.0 &13.2  &45.2    &74.0    &-  &- &14.6  \\
        X-Pool \cite{DBLP:conf/cvpr/X-PoolGort}         &V+T&46.9    &72.8    &82.2  &2.0 &14.3  &44.6    &73.2    &82.0  &2.0 &15.4  \\
        TS2-Net \cite{liu_ts2-net_2022}        &V+T&47.0    &74.5    &83.8  &2.0 &13.0  &41.8    &71.6    &82.0  &2.0 &14.8  \\
        TEFAL \cite{audio_enhanced}          &A+V+T&49.4    &75.9    &83.9  &2.0 &12.0  &-    &-    &-  &- &-  \\
        CLIP-ViP \cite{xue_clip-vip_2023}       &V+T&50.1    &74.8    &84.6  &1.0 &-     &48.6    &77.1    &84.4  &2.0 &-  \\
        AVIGATE \cite{DBLP:conf/cvpr/JeongPKK25}    &A+V+T &50.2 &74.3 &83.2 & - & -     &-  &-  &-  &-  &-  \\
        T-MASS \cite{wang_text_2024}         &V+T&50.2    &75.3    &85.1  &1.0 &11.9 &50.9    &77.2    &85.3  &1.0 &12.1   \\
        \rowcolor{gray!20}
        \textbf{GAIS}(Ours)           &A+V+T&\textbf{57.0}    &\textbf{83.1}    & \textbf{90.9}& \textbf{1.0} & \textbf{7.6} &\textbf{54.3}   &\textbf{79.8}    &\textbf{87.6}  &\textbf{1.0} &\textbf{10.0}  \\
        \midrule
        \textit{\textcolor{OliveGreen}{ViT-B/16}}  &&\multicolumn{5}{l|}{} &\multicolumn{5}{l}{} \\
        X-Pool \cite{DBLP:conf/cvpr/X-PoolGort}         &V+T&48.2    &73.7    &82.6  &2.0 &12.7  &47.3    &74.8    &82.8  &2.0 &14.2  \\
        HunYuan \cite{hunyuan}  &V+T& 49.7 & 75.0 & 83.5 & 2.0 & 11.4 & 45.0 & 75.6 & 83.4 & 2.0 & 12.0 \\
        TEFAL \cite{audio_enhanced}         &A+V+T&49.9    &76.2    &85.4  &1.0 &11.4  &-    &-    &-  &- &-  \\
        AVIGATE \cite{DBLP:conf/cvpr/JeongPKK25}       &A+V+T&52.1  &76.4    &85.2  & - & -     &-  &-  &-  &-  &-  \\
        T-MASS \cite{wang_text_2024}        &V+T&52.7    &77.1    &85.6  &1.0 &10.5 &53.3    &80.1    &87.7  &1.0 &9.8 \\
        CLIP-ViP \cite{xue_clip-vip_2023}      &V+T&54.2    &77.2    &84.8  &1.0 &-  &50.5    &78.4    &87.1  &1.0 &-  \\
        $\mathrm{VALOR_L}^*$ \cite{DBLP:journals/pami/valor} & A+V+T & 54.4 & 79.8 & 87.6 & 1.0 & - & 56.6 & 83.3 & 88.8 & 1.0 & - \\
        ExCae \cite{DBLP:journals/corr/excae} & V+T & 55.0 & 84.6 & 91.3 & 1.0 & 6.0 & 53.9 & 80.6 & 87.3 & 1.0 & 9.9 \\
        $\mathrm{InternVid^*}$ \cite{DBLP:conf/iclr/InternVid}  &V+T& 55.0 & - & - & - & - & 51.7 & - & - & - & - \\
        \rowcolor{gray!20}
        \textbf{GAIS}(Ours)           &A+V+T& 58.9    &84.6    & 94.0  &1.0 & 5.2 &57.6     &81.7    &89.4  &1.0 &9.9 \\
        $\mathrm{VALOR_L}^*$ \cite{DBLP:journals/pami/valor}+DSL  &A+V+T& 59.9 & 83.5 & 89.6 & - & - & 61.5 & 85.3 & 90.4 & - & - \\
        \rowcolor{gray!20}
        \textbf{GAIS}(Ours)+DSL   &A+V+T& \textbf{68.0}    &\textbf{89.0}    &\textbf{94.0}  &\textbf{1.0}    &\textbf{4.4} &\textbf{64.1}   &\textbf{86.0}    &\textbf{91.0}  &\textbf{1.0} &\textbf{9.4} \\
        \bottomrule

    \end{tabular}

    \caption{Text-to-video comparisons on MSR-VTT 9k split and DiDeMo. V, A, T denote Video, Audio, and Text modalities, respectively. Both ViT-B/32 and ViT-B/16 backbones are adopted for evaluation. Bold denotes the best performance. "-": result is unavailable. It is worth noting that the methods marked with * use larger visual encoder (\textit{e.g.}, ViT-L/14)}
    \label{tab:overall_result}
\end{table*}

\begin{table}[h]
\fontsize{9}{10}\selectfont
    \setlength{\tabcolsep}{1.5mm}
    \centering
    \begin{tabular}{l|l l l l l}
        \toprule
        Method & R@1$\uparrow$ & R@5$\uparrow$ & R@10$\uparrow$ & MdR$\downarrow$ & MnR$\downarrow$ \\
         \midrule
         ECLIPSE \cite{eccv_EclipSE}             &57.8   &88.4    &94.3   &1.0    &4.3 \\
         X-Pool \cite{DBLP:conf/cvpr/X-PoolGort} &60.0   &90.0    &95.0   &1.0    &3.8 \\
         TEFAL \cite{audio_enhanced}             &61.0   &90.4    &95.3   &1.0    &3.8    \\
         UATVR \cite{DBLP:conf/iccv/UATVR}       &61.3   &91.0    &95.6   &1.0    &3.3  \\
         T-MASS \cite{wang_text_2024}            &63.0   &92.3    &96.4   &1.0    &3.2 \\
         AVIGATE \cite{DBLP:conf/cvpr/JeongPKK25} &63.1  &90.7    &95.5   &1.0    & - \\
         Cap4Video \cite{wu_cap4video_2023}      &66.6   &\textbf{93.1}    &\textbf{97.0}   &1.0     & 2.7 \\
         \rowcolor{gray!20}
         \textbf{GAIS}(Ours) &\textbf{68.1}   &\textbf{93.1}    &\textbf{97.0}&\textbf{1.0}    &\textbf{2.4} \\
        \bottomrule
    \end{tabular}
    
    \caption{Text-to-Video results on VATEX with ViT-B/32.}
    \label{tab:vatex_result}
\end{table}

\begin{table}[b]
\fontsize{9}{10}\selectfont
    \setlength{\tabcolsep}{1.5mm}
    \centering
    \begin{tabular}{l|l l l l l}
        \toprule
        Method & R@1$\uparrow$ & R@5$\uparrow$ & R@10$\uparrow$ & MdR$\downarrow$ & MnR$\downarrow$ \\
         \midrule
         CLIP4Clip \cite{luo_clip4clip_2021} &22.6   &41.0    &49.1   &11.0    &61.0 \\
         DRL \cite{DBLP:journals/corr/drl-wang} &24.9   &45.7    &55.3   &7.0    &- \\
         X-Pool \cite{DBLP:conf/cvpr/X-PoolGort} &25.2   &43.7    &53.5   &8.0    &53.2 \\
         DiffusionRet \cite{DBLP:conf/iccv/DiffusionRet} &25.2   &43.7    &53.5   &8.0    &40.7 \\
         TEFAL \cite{audio_enhanced} &26.8    &46.1   &56.5   &7.0    &44.4    \\
         CLIP-ViP \cite{xue_clip-vip_2023} &25.6   &45.3    &54.4   &8.0    &-  \\
         T-MASS \cite{wang_text_2024} &28.9   &48.2   &57.6   &6.0    &43.3 \\
         \rowcolor{gray!20}
         \textbf{GAIS}(Ours) &\textbf{30.9}   &\textbf{50.8}    &\textbf{60.3}   &\textbf{5.0}    &\textbf{37.2} \\
        \bottomrule
    \end{tabular}
    
    \caption{Text-to-Video results on LSMDC with ViT-B/32.}
    \label{tab:lsmdc_result}
    
\end{table}

\begin{table}
\fontsize{9}{10}\selectfont
    \centering
    \setlength{\tabcolsep}{1.5mm}
    \begin{tabular}{l|l l l l l}
        \toprule
        Method & R@1$\uparrow$ & R@5$\uparrow$ & R@10$\uparrow$ & MdR$\downarrow$ & MnR$\downarrow$ \\
         \midrule
         CLIP4Clip \cite{luo_clip4clip_2021}& 42.7 & 70.9 & 80.6 & 2.0 & 11.6 \\
         CenterCLIP \cite{DBLP:conf/sigir/centerclip} & 42.8 & 71.7 & 82.2 & 2.0 & 10.9 \\
         X-Pool \cite{DBLP:conf/cvpr/X-PoolGort} &44.4   &73.3    &84.0   &2.0    &9.0 \\
         TS2-Net \cite{liu_ts2-net_2022} & 45.3 & 74.1 & 83.7 & 2.0 & 9.2 \\
         DiffusionRet \cite{DBLP:conf/iccv/DiffusionRet} &47.7   &73.8    &84.5   &2.0    &8.8 \\
         UATVR \cite{DBLP:conf/iccv/UATVR} &46.9    &73.8   &83.8   &2.0    &8.6    \\
         T-MASS \cite{wang_text_2024} &47.7   &78.0   &86.3   &2.0    &8.0 \\
         AVIGATE \cite{DBLP:conf/cvpr/JeongPKK25} &49.7 &75.3 &83.7 &- &- \\
         \rowcolor{gray!20}
         \textbf{GAIS}(Ours) &\textbf{56.2}   &\textbf{84.0}   &\textbf{91.4}   &\textbf{1.0}    &\textbf{6.1} \\
        \bottomrule
    \end{tabular}
    
    \caption{Video-to-Text results on MSR-VTT 9k split.}
    \label{tab:v2t_result}
\end{table}

\subsection{Performance Comparison}
We conducted comparative experiments with previous methods on the MSR-VTT, DiDeMo, VATEX, and LSMDC with results presented in \cref{tab:overall_result,tab:vatex_result,tab:lsmdc_result,tab:v2t_result}. 

On MSR-VTT, GAIS surpasses both audio-aware methods (\textit{e.g.}, AVIGATE \cite{DBLP:conf/cvpr/JeongPKK25}) and audio-agnostic methods (\textit{e.g.}, T-MASS \cite{wang_text_2024}, InternVid \cite{DBLP:conf/iclr/InternVid}) under both ViT-B/32 and ViT-B/16 backbones. Using ViT-B/32, GAIS achieves absolute gains of \textbf{6.8\%} R@1 and \textbf{7.8\%} R@5 over the best prior method; ViT-B/16 further improves R@1 by an additional 1.9\%. Even compared with VALOR \cite{DBLP:journals/pami/valor} and InernVid with larger backbone or enhanced by DSL post-processing \cite{DBLP:journals/corr/dual_softmax}, GAIS remains superior.
It is worth noting that CLIP-ViP \cite{xue_clip-vip_2023}, which augments CLIP with rich frame-level textual descriptions instead of audio cues, also achieves strong performance. However, GAIS still outperforms CLIP-ViP across all metrics. These results indicate that fine-grained audio grounding contributes more than global fusion, and variance-scaled perturbation provides gains orthogonal to visual-text alignment. More retrieval examples and failure cases in Appendix~\ref{sec:more_quali_result}.

On DiDeMo, GAIS improves R@1 by \textbf{3.4\%} and yields consistent gains across R@5 and R@10. Similar trends are observed on VATEX and LSMDC (both evaluated with ViT-B/32), where GAIS achieves \textbf{+1.5\%} and \textbf{+2.0\%} R@1 improvements, respectively, over the strongest baselines.

For video-to-text retrieval (\cref{tab:v2t_result}), GAIS also outperforms prior SOTA methods across all metrics. Relative to the audio-enhanced AVIGATE, GAIS delivers gains of \textbf{6.5\%} R@1, \textbf{8.7\%} R@5, and \textbf{7.7\%} R@10, underscoring the effectiveness of combining fine-grained audio-visual fusion with structure-aware perturbation for bidirectional retrieval. Additional comparisons with other DSL-enhanced methods, as well as the corresponding improvements, are provided in Appendix~\ref{sec:more_quanti_result}.

\subsection{Ablation on FGF and SVSP}
To further evaluate the effectiveness of different components of the model, we conduct additional experiments base on the ViT-B/32 backbone.

\textbf{Main Ablation Study.} We first provide a step-by-step ablation, showing that FGF and SVSP contribute complementary gains and the full GAIS model achieves the best performance. \cref{tab:ablation_analysis} presents the incremental ablation analysis from the baseline to the full GAIS model. Introducing FGF alone yields a large improvement (R@1: 48.7$\rightarrow$53.8). Incorporating SVSP alone also improves performance over the baseline (R@1: 48.7$\rightarrow$50.3). On top of the baseline, FGF improves R@1 by \textbf{+5.1} while SVSP adds another \textbf{+3.2}, confirming their complementary benefits. Combining both modules achieves the best performance (R@1: 57.0).

\textbf{FGF Ablation.}\label{expri:fusion} \cref{tab:fusion_method} compares different audio–visual fusion strategies.
Methods that either omit audio entirely (“No Fusion”) or combine features at the sample level simply broadcast audio cues uniformly, yielding limited improvement. Concat-based fusion and standard Cross-Attention introduce feature interactions but still lack temporal selectivity, causing irrelevant audio segments to dilute meaningful cues. Cross-Attention with text guidance improves alignment but remains frame-agnostic and treats all timestamps equally.

Removing text guidance in our fusion module (“FGF w/o T”) results in a noticeable drop in performance, indicating that textual relevance is essential for identifying which audio frames contribute useful semantic information.
In contrast, our full FGF module adaptively selects salient audio–visual evidence on a frame level, conditioned on the linguistic query. This allows the model to suppress redundant ambient sounds and emphasize semantically meaningful audio segments, leading to consistent gains across R@1, R@5, and MnR. 

\begin{figure}[ht]
    \centering
    \includegraphics[width=1\linewidth]{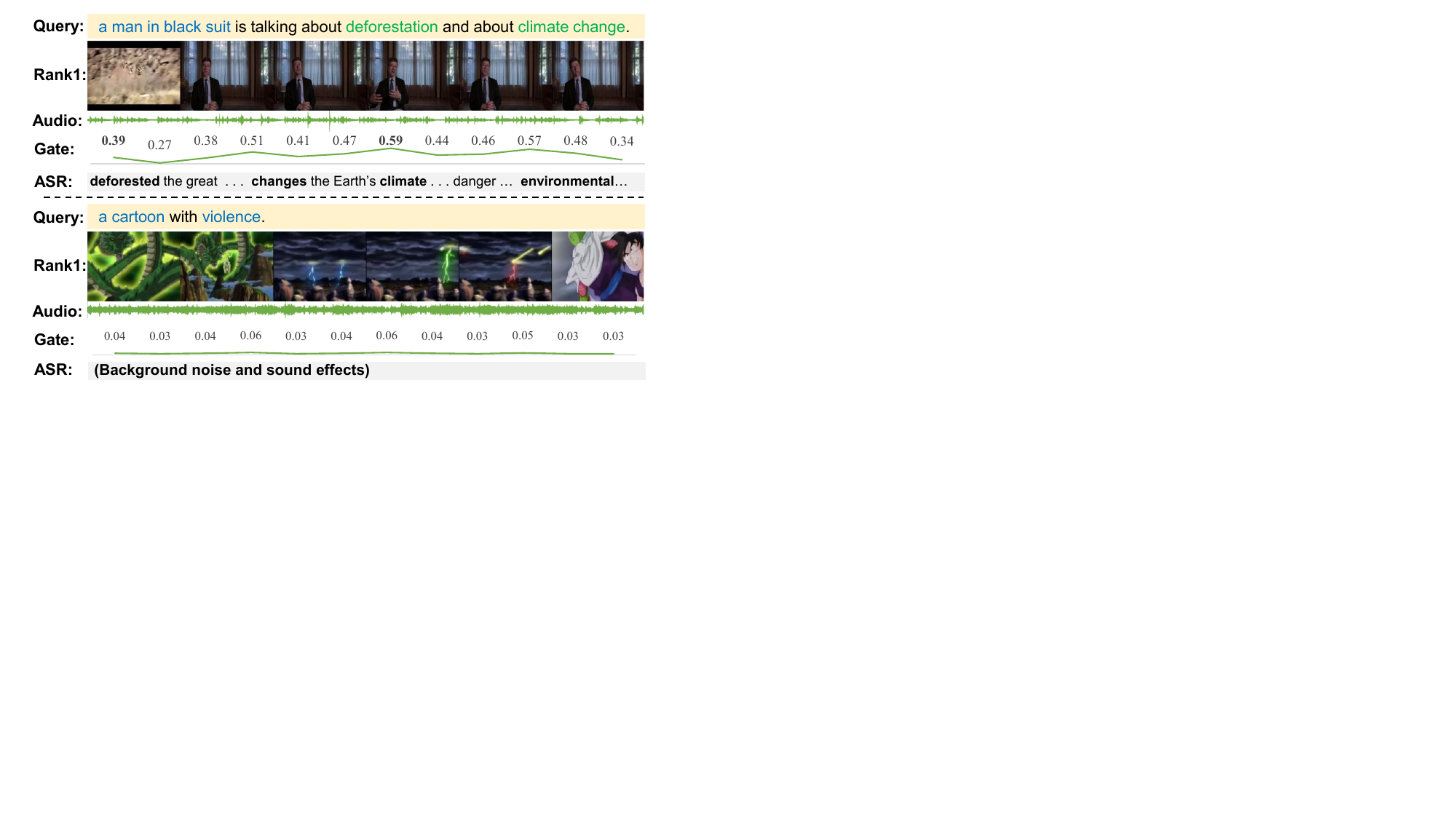}
    \caption{Text-guided frame-level audio–visual gating. FGF assigns high gate values to frames whose audio signals align with the query semantics, and suppresses uninformative or noisy segments, illustrating its fine-grained and interpretable fusion behavior. }
    \label{fig:gate_vis}
\end{figure}

To further illustrate how FGF selectively leverages audio signals, we visualize the frame-level gating behavior in \cref{fig:gate_vis}. As shown in the top example, when the query describes spoken content, the gating module assigns high weights to frames containing clear dialogue, effectively aligning salient audio cues with the textual semantics. In contrast, in the bottom example where the video contains only background music and sound effects, the gating scores remain consistently low, indicating that FGF successfully suppresses uninformative audio. These qualitative results demonstrate that FGF not only improves retrieval accuracy but also provides interpretable, text-guided control over audio–visual fusion. 

These results verify that not all audio frames are equally informative, and effective multimodal fusion requires text-conditioned temporal selection, rather than uniform or global mixing.

\begin{table}[t]
\fontsize{9}{10}\selectfont
    \centering
    \setlength{\tabcolsep}{1.5mm}
    \begin{tabular}{l|cccccc}
    \toprule
        Model Variant  & FGF & SVSP & R@1 & R@5 & R@10 & MnR \\
        \midrule
        Baseline & \ding{55} & \ding{55} & 48.7& 75.5& 85.1 & 11.7\\
        + FGF only & \ding{51} & \ding{55}  & 53.8& 81.1& 88.5 & 8.0\\
        + SVSP only & \ding{55} & \ding{51} &  50.3& 77.0&  85.8 & 10.6\\
        \rowcolor{gray!20}
        GAIS(Ours) & \ding{51} & \ding{51} & 57.0 & 83.1 & 90.9 & 7.6 \\
        \bottomrule
    \end{tabular}
    \caption{Incremental ablation analysis from baseline to the full GAIS model.}
    \label{tab:ablation_analysis}
\end{table}

\begin{table}[t]
\fontsize{9}{11}\selectfont
    \centering
    \setlength{\tabcolsep}{1mm}
    \begin{tabular}{l|ccccc}
        \toprule
        Fusion Strategy  & R@1$\uparrow$ & R@5$\uparrow$ & R@10$\uparrow$ & MdR$\downarrow$ & MnR$\downarrow$ \\
        \midrule
        No Fusion           & 52.0       & 78.6        & 86.8                & 1.0          & 10.5       \\
        Sample-level       &52.0    &79.0    &87.9    &1.0        &10.2    \\
        CrossAttn \cite{eccv_EclipSE} & 51.7 & 78.4& 85.8 & 1.0 & 10.1 \\
        ConcatMLP & 52.6&79.0 &86.9 &1.0 &  10.4\\
        CrossAttn w/ T \cite{audio_enhanced}& 53.2 & 79.3&88.0 &1.0 &10.2 \\
        FGF w/o T & 52.5 & 79.4&87.6 &1.0 &10.2 \\
        \rowcolor{gray!20}
        \textbf{FGF(Ours)}    &\textbf{55.0}&\textbf{83.0}&\textbf{89.9}&\textbf{1.0} &\textbf{7.7}\\
        
        \bottomrule
    \end{tabular}
    \caption{FGF achieves the best retrieval accuracy by integrating audio cues selectively under textual guidance, highlighting the importance of fine-grained and context-aware multimodal alignment.}
    \label{tab:fusion_method}
\end{table}

\textbf{Ablation on Perturbation Mechanisms.}\label{experiment:ast} \cref{tab:SVSP_comparison} compares different perturbation strategies. Removing perturbation (“No Perturb”) leads to noticeably worse performance, showing that allowing controlled flexibility in the text embedding is beneficial for modeling semantic ambiguity in video descriptions. Applying perturbation with random magnitude (“STP”) yields only marginal improvement and even degrades MnR, indicating that uncontrolled noise may distort semantic meaning and produce unstable similarity estimates. Moreover, STP requires multiple sampling passes during inference to obtain reliable rankings, which significantly increases computational cost (98.2s vs. 6.1s).

\begin{figure}[t]
    \centering
    \includegraphics[width=1\linewidth]{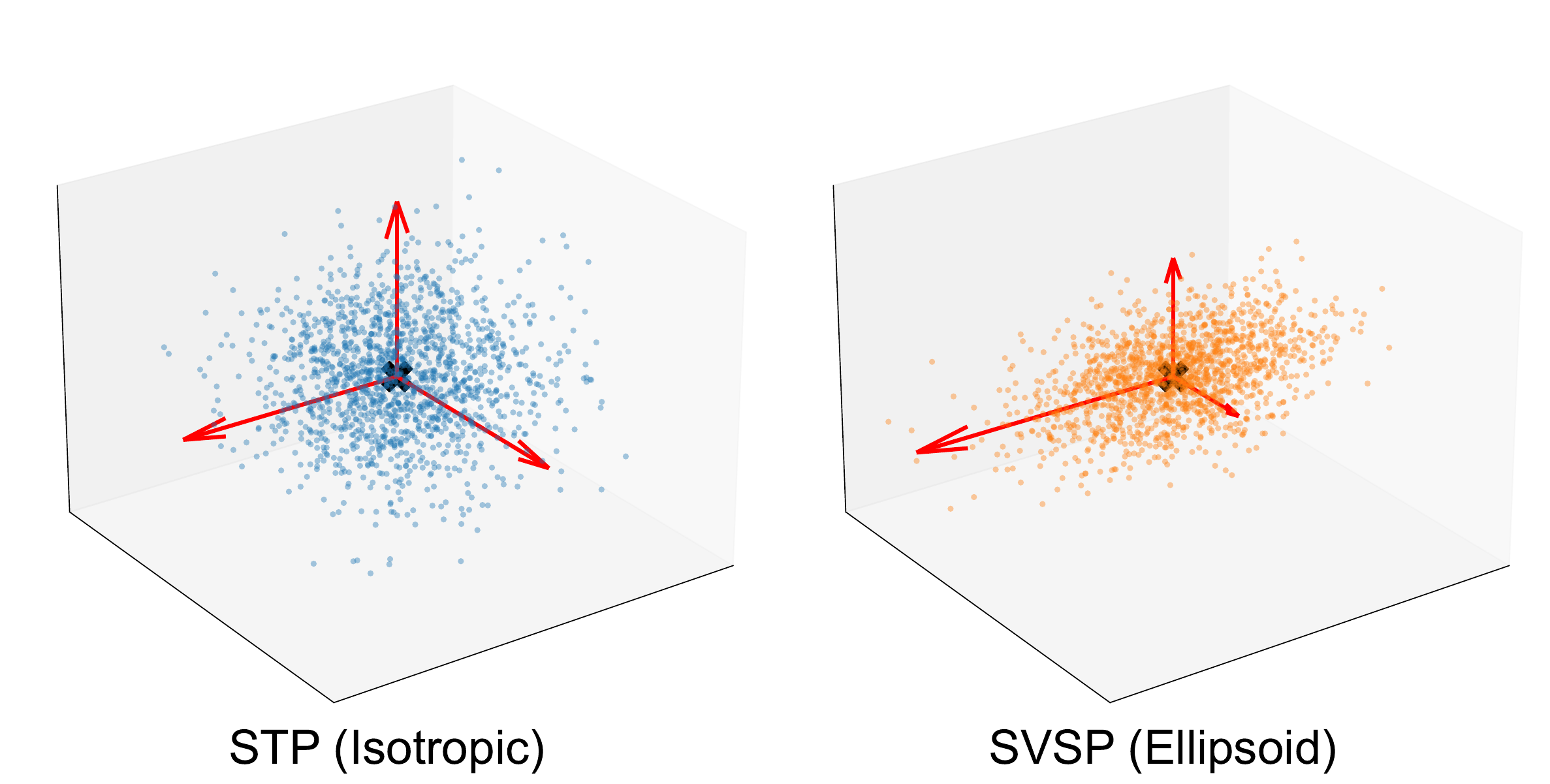}
    \caption{Comparison of perturbation distributions in the text embedding space.
    Left: Fixed-magnitude stochastic perturbation produces a dispersed isotropic distribution.Right: SVSP scales perturbation by semantic variance, forming a compact ellipsoidal neighborhood around the original embedding (black ‘X’). }
    \label{fig:SVSP_vis}
\end{figure}

In contrast, SVSP achieves the best retrieval accuracy while maintaining single-pass inference efficiency. By scaling perturbation per semantic dimension according to cross-modal variance, SVSP selectively adjusts only the embedding components associated with semantic uncertainty, while leaving grounded components unchanged. This leads to a compact and coherent embedding neighborhood, improving both robustness during training and stability during inference. As shown in \cref{fig:SVSP_vis}, STP results in a dispersed isotropic perturbation cloud, whereas SVSP produces a centered ellipsoid-shaped distribution that preserves semantic consistency. In addition to the quantitative gains, SVSP also produces significantly more stable similarity distributions across queries. A detailed comparison between stochastic perturbation (STP) and SVSP is provided in Appendix~\ref{sec:more_quali_result}.

Overall, these results demonstrate that the key factor is not whether perturbation is applied, but whether the magnitude of perturbation is guided by semantic variance.

\begin{table}[]
\fontsize{9}{11}\selectfont
    \centering
    \begin{tabular}{l|cccc|c}
    \toprule
     Methods & R@1 & R@5 & R@10 & MnR & Time Cost \\
    \midrule
    No Perturb & 53.8 & 80.4 & 88.2 & 8.25 & \textbf{6.1s} \\
     STP & 54.0 & 80.6 & 89.2 & \textbf{7.4} & 98.2s \\
     \rowcolor{gray!20}
     SVSP(Ours) & \textbf{57.0} & \textbf{83.1} & \textbf{90.9} & 7.6 & 6.5s \\
    \bottomrule
    \end{tabular}
    
    \caption{Ablation of perturbation mechanisms. SVSP improves retrieval accuracy while maintaining single-pass inference. Random perturbation yields unstable ranking and significantly higher inference cost. }
    \label{tab:SVSP_comparison}
\end{table}

\subsection{Further Analysis}

\textbf{Audio Encoder Selection.} To study the influence of the audio encoder, we evaluate Whisper \cite{DBLP:conf/icml/RadfordKXBMS23} and Wav2vec2.0~\cite{DBLP:conf/nips/BaevskiZMA20} on the DiDeMo dataset with different model sizes. As shown in \cref{tab:audio_compare}, all variants achieve comparable top-1 accuracy (R@1 $\approx$ 54\%), indicating that GAIS benefits consistently from strong audio features regardless of whether the encoder is trained for ASR (Whisper) or self-supervised speech modeling (Wav2vec2.0).

Scaling the audio encoder yields limited gains. For Whisper, moving from base (74M) to small (244M) results in only marginal improvement in R@1 (+0.5) and R@10 (+1.5), with slightly worse MnR. A similar trend appears for Wav2vec2.0, where the large model (317M) raises R@1 to 55.0 but does not consistently improve R@5 or MnR. Given the small performance differences and the additional computational cost of larger encoders, we adopt Whisper-base as the default audio encoder for all experiments.

\begin{table}
\fontsize{9}{11}\selectfont
    \centering
    \setlength{\tabcolsep}{1mm}
    \begin{tabular}{l|c|c|ccccc}
        \toprule
        Audio Encoder & Size & Parameters  & R@1 & R@5 & R@10 & MnR \\
        \midrule
        \multirow{2}{*}{Whisper}    & base   & 74M       & 53.5   &77.8    &85.8   & 10.9   \\
            & small  & 244M      &54.0    &78.6    &\textbf{87.3}     &11.0    \\
        \midrule
        \multirow{2}{*}{Wav2vec2.0} & base   & 95M       &54.0    &\textbf{80.0}    &87.1  &\textbf{10.6}   \\
         & large   & 317M       &\textbf{55.0}    &79.9    &86.3  &10.9    \\
        \bottomrule
    \end{tabular}
    \caption{Text-to-video comparisons on DiDeMo across audio encoder variant. Whisper-base is used by default.}
    \label{tab:audio_compare}
\end{table}

\begin{table}[]
\fontsize{9}{11}\selectfont
    \centering
    \begin{tabular}{l|p{1cm}p{1.2cm}cp{1cm}}
    \toprule
    Method & Total Params(B) & Trainable Params(B) & GFLOPs & Inference Time(ms) \\
    \midrule
    AVIGATE \cite{DBLP:conf/cvpr/JeongPKK25}&283.6M& 169.5M & 139.60 & 36.1 \\
    T-MASS \cite{wang_text_2024} & 152.6M & 127.2M & 52.97 & 24.5\\
    GAIS(Ours)   & 227.9M& 127.2M & 61.97 & 20.0 \\
    \bottomrule
    \end{tabular}
    \caption{Comparison of Model Complexity and Efficiency.}
    \label{tab:efficiency}
\end{table}

\begin{figure}[t]
    \centering
    \includegraphics[width=0.9\linewidth]{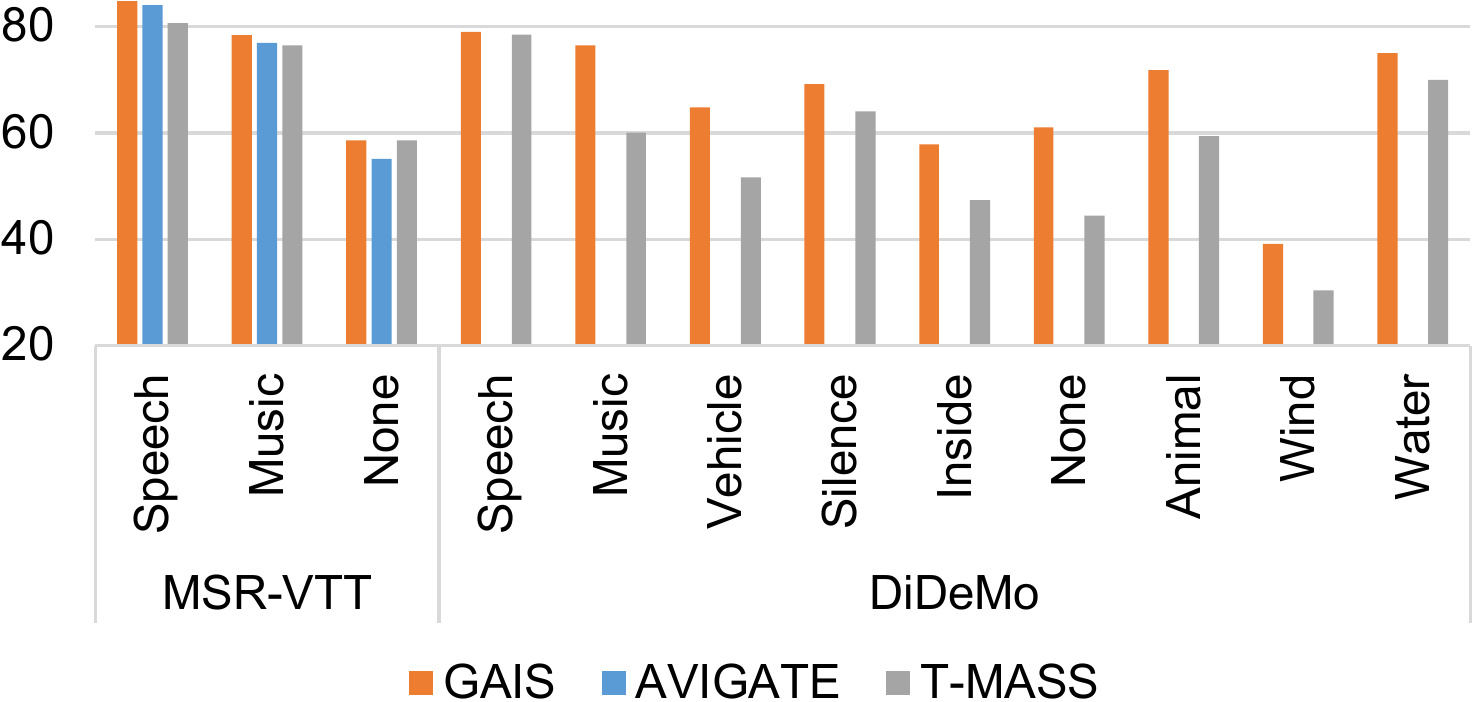}
    \caption{Per-audio-category retrieval performance (R@1) on MSR-VTT and DiDeMo. The audio classification was performed by the YAMNet model \cite{gemmeke2017audioset}.}
    \label{fig:audio_bucket}
\end{figure}

\textbf{Audio Bucket Comparison.} We group videos by their dominant audio type and report R@1 within each bucket (only categories with $\ge$20 samples, see \cref{fig:audio_bucket}). As shown in the treemap statistics \cref{fig:audio_treemap} in the supplementary material, Speech and Music together account for over $60\%$ of videos in both MSR-VTT and DiDeMo, while the remaining categories form a long-tail distribution. GAIS shows clear improvements in these major buckets, indicating that text-guided audio–visual fusion is most beneficial when audio cues carry meaningful semantic information.


\textbf{Efficiency Comparison.} As shown in \cref{tab:efficiency}, we include the parameters and FLOPs of the Whisper-base audio encoder for all audio-aware methods. Despite this addition, GAIS still uses fewer total parameters than AVIGATE (227.9M vs. 283.6M) and requires less than half of its FLOPs. GAIS also achieves the fastest inference time (20.0 ms), showing that our design improves retrieval accuracy while remaining highly efficient.

\section{Conclusion}
In this work, we introduced GAIS, a text-to-video retrieval framework that integrates frame-level gated audio–visual fusion with semantic variance-scaled text perturbation. The gating mechanism captures fine-grained multimodal dependencies, while SVSP enhances the robustness and discriminability of textual embeddings with single-pass inference. Experiments on four benchmarks demonstrate strong performance and improved interpretability through dynamic modality weighting.

\textbf{Applicability and limitations.} GAIS is most effective when audio carries meaningful, query-relevant cues, particularly in short or medium-length videos where informative audio segments appear sparsely. Its benefits diminish in silent, noisy, or weakly audio-correlated videos, where gating provides limited additional signal. Moreover, GAIS currently depends on global text–video interaction, making it incompatible with two-stage re-ranking frameworks. Extending GAIS to long-form and re-ranking–friendly settings is a promising direction.

\small
\bibliographystyle{ieeenat_fullname}
\bibliography{main}

@String(CVPR= {IEEE Conf. Comput. Vis. Pattern Recog.})

@String(ICCV= {Int. Conf. Comput. Vis.})

@String(ECCV= {Eur. Conf. Comput. Vis.})

@String(ICASSP=	{ICASSP})

@String(ICLR = {Int. Conf. Learn. Represent.})

@String(CVPR  = {CVPR})

@String(ICCV  = {ICCV})

@String(ECCV  = {ECCV})

@String(ICLR  = {ICLR})

@inproceedings{liu_ts2-net_2022,
  author       = {Yuqi Liu and
                  Pengfei Xiong and
                  Luhui Xu and
                  Shengming Cao and
                  Qin Jin},
  editor       = {Shai Avidan and
                  Gabriel J. Brostow and
                  Moustapha Ciss{\'{e}} and
                  Giovanni Maria Farinella and
                  Tal Hassner},
  title        = {TS2-Net: Token Shift and Selection Transformer for Text-Video Retrieval},
  booktitle    = {Computer Vision - {ECCV} 2022 - 17th European Conference, Tel Aviv,
                  Israel, October 23-27, 2022, Proceedings, Part {XIV}},
  series       = {Lecture Notes in Computer Science},
  volume       = {13674},
  pages        = {319--335},
  publisher    = {Springer},
  year         = {2022},
  url          = {https://doi.org/10.1007/978-3-031-19781-9\_19},
  doi          = {10.1007/978-3-031-19781-9\_19},
  bibsource    = {dblp computer science bibliography, https://dblp.org}
}

@inproceedings{xue_clip-vip_2023,
  title={Clip-vip: Adapting pre-trained image-text model to video-language alignment},
  author={Xue, Hongwei and Sun, Yuchong and Liu, Bei and Fu, Jianlong and Song, Ruihua and Li, Houqiang and Luo, Jiebo},
  booktitle={The Eleventh International Conference on Learning Representations},
  year={2022}
}

@article{luo_clip4clip_2021,
  author       = {Huaishao Luo and
                  Lei Ji and
                  Ming Zhong and
                  Yang Chen and
                  Wen Lei and
                  Nan Duan and
                  Tianrui Li},
  title        = {CLIP4Clip: An empirical study of {CLIP} for end to end video clip
                  retrieval and captioning},
  journal      = {Neurocomputing},
  volume       = {508},
  pages        = {293--304},
  year         = {2022},
  url          = {https://doi.org/10.1016/j.neucom.2022.07.028},
  doi          = {10.1016/J.NEUCOM.2022.07.028},
  bibsource    = {dblp computer science bibliography, https://dblp.org}
}

@inproceedings{wang_text_2024,
  author       = {Jiamian Wang and
                  Pichao Wang and
                  Guohao Sun and
                  Dongfang Liu and
                  Sohail A. Dianat and
                  Raghuveer Rao and
                  Majid Rabbani and
                  Zhiqiang Tao},
  title        = {Text Is {MASS:} Modeling as Stochastic Embedding for Text-Video Retrieval},
  booktitle    = {{IEEE/CVF} Conference on Computer Vision and Pattern Recognition,
                  {CVPR} 2024, Seattle, WA, USA, June 16-22, 2024},
  pages        = {16551--16560},
  publisher    = {{IEEE}},
  year         = {2024},
  url          = {https://doi.org/10.1109/CVPR52733.2024.01566},
  doi          = {10.1109/CVPR52733.2024.01566},
  bibsource    = {dblp computer science bibliography, https://dblp.org}
}

@inproceedings{bain_frozen_2021,
  author       = {Max Bain and
                  Arsha Nagrani and
                  G{\"{u}}l Varol and
                  Andrew Zisserman},
  title        = {Frozen in Time: {A} Joint Video and Image Encoder for End-to-End Retrieval},
  booktitle    = {2021 {IEEE/CVF} International Conference on Computer Vision, {ICCV}
                  2021, Montreal, QC, Canada, October 10-17, 2021},
  pages        = {1708--1718},
  publisher    = {{IEEE}},
  year         = {2021},
  url          = {https://doi.org/10.1109/ICCV48922.2021.00175},
  doi          = {10.1109/ICCV48922.2021.00175},
  bibsource    = {dblp computer science bibliography, https://dblp.org}
}

@inproceedings{wu_cap4video_2023,
      author       = {Wenhao Wu and
                      Haipeng Luo and
                      Bo Fang and
                      Jingdong Wang and
                      Wanli Ouyang},
      title        = {Cap4Video: What Can Auxiliary Captions Do for Text-Video Retrieval?},
      booktitle    = {{IEEE/CVF} Conference on Computer Vision and Pattern Recognition,
                      {CVPR} 2023, Vancouver, BC, Canada, June 17-24, 2023},
      pages        = {10704--10713},
      publisher    = {{IEEE}},
      year         = {2023},
      url          = {https://doi.org/10.1109/CVPR52729.2023.01031},
      doi          = {10.1109/CVPR52729.2023.01031},
      bibsource    = {dblp computer science bibliography, https://dblp.org}
}

@inproceedings{DBLP:conf/cvpr/ChenZJW20,
  author       = {Shizhe Chen and
                  Yida Zhao and
                  Qin Jin and
                  Qi Wu},
  title        = {Fine-Grained Video-Text Retrieval With Hierarchical Graph Reasoning},
  booktitle    = {2020 {IEEE/CVF} Conference on Computer Vision and Pattern Recognition,
                  {CVPR} 2020, Seattle, WA, USA, June 13-19, 2020},
  pages        = {10635--10644},
  publisher    = {Computer Vision Foundation / {IEEE}},
  year         = {2020},
  url          = {https://openaccess.thecvf.com/content\_CVPR\_2020/html/Chen\_Fine-Grained\_Video-Text\_Retrieval\_With\_Hierarchical\_Graph\_Reasoning\_CVPR\_2020\_paper.html},
  doi          = {10.1109/CVPR42600.2020.01065},
  bibsource    = {dblp computer science bibliography, https://dblp.org}
}

@inproceedings{DBLP:conf/mm/WuHTLL21,
  author       = {Peng Wu and
                  Xiangteng He and
                  Mingqian Tang and
                  Yiliang Lv and
                  Jing Liu},
  editor       = {Heng Tao Shen and
                  Yueting Zhuang and
                  John R. Smith and
                  Yang Yang and
                  Pablo C{\'{e}}sar and
                  Florian Metze and
                  Balakrishnan Prabhakaran},
  title        = {HANet: Hierarchical Alignment Networks for Video-Text Retrieval},
  booktitle    = {{MM} '21: {ACM} Multimedia Conference, Virtual Event, China, October
                  20 - 24, 2021},
  pages        = {3518--3527},
  publisher    = {{ACM}},
  year         = {2021},
  url          = {https://doi.org/10.1145/3474085.3475515},
  doi          = {10.1145/3474085.3475515},
  bibsource    = {dblp computer science bibliography, https://dblp.org}
}

@inproceedings{DBLP:conf/cvpr/WangZ021,
  author       = {Xiaohan Wang and
                  Linchao Zhu and
                  Yi Yang},
  title        = {{T2VLAD:} Global-Local Sequence Alignment for Text-Video Retrieval},
  booktitle    = {{IEEE} Conference on Computer Vision and Pattern Recognition, {CVPR}
                  2021, virtual, June 19-25, 2021},
  pages        = {5079--5088},
  publisher    = {Computer Vision Foundation / {IEEE}},
  year         = {2021},
  url          = {https://openaccess.thecvf.com/content/CVPR2021/html/Wang\_T2VLAD\_Global-Local\_Sequence\_Alignment\_for\_Text-Video\_Retrieval\_CVPR\_2021\_paper.html},
  doi          = {10.1109/CVPR46437.2021.00504},
  bibsource    = {dblp computer science bibliography, https://dblp.org}
}

@inproceedings{DBLP:conf/eccv/Gabeur0AS20,
  author       = {Valentin Gabeur and
                  Chen Sun and
                  Karteek Alahari and
                  Cordelia Schmid},
  editor       = {Andrea Vedaldi and
                  Horst Bischof and
                  Thomas Brox and
                  Jan{-}Michael Frahm},
  title        = {Multi-modal Transformer for Video Retrieval},
  booktitle    = {Computer Vision - {ECCV} 2020 - 16th European Conference, Glasgow,
                  UK, August 23-28, 2020, Proceedings, Part {IV}},
  series       = {Lecture Notes in Computer Science},
  volume       = {12349},
  pages        = {214--229},
  publisher    = {Springer},
  year         = {2020},
  url          = {https://doi.org/10.1007/978-3-030-58548-8\_13},
  doi          = {10.1007/978-3-030-58548-8\_13},
  bibsource    = {dblp computer science bibliography, https://dblp.org}
}

@inproceedings{DBLP:conf/cvpr/LeiLZGBB021,
  author       = {Jie Lei and
                  Linjie Li and
                  Luowei Zhou and
                  Zhe Gan and
                  Tamara L. Berg and
                  Mohit Bansal and
                  Jingjing Liu},
  title        = {Less Is More: ClipBERT for Video-and-Language Learning via Sparse
                  Sampling},
  booktitle    = {{IEEE} Conference on Computer Vision and Pattern Recognition, {CVPR}
                  2021, virtual, June 19-25, 2021},
  pages        = {7331--7341},
  publisher    = {Computer Vision Foundation / {IEEE}},
  year         = {2021},
  url          = {https://openaccess.thecvf.com/content/CVPR2021/html/Lei\_Less\_Is\_More\_ClipBERT\_for\_Video-and-Language\_Learning\_via\_Sparse\_Sampling\_CVPR\_2021\_paper.html},
  doi          = {10.1109/CVPR46437.2021.00725},
  bibsource    = {dblp computer science bibliography, https://dblp.org}
}

@inproceedings{DBLP:conf/icml/RadfordKHRGASAM21,
  author       = {Alec Radford and
                  Jong Wook Kim and
                  Chris Hallacy and
                  Aditya Ramesh and
                  Gabriel Goh and
                  Sandhini Agarwal and
                  Girish Sastry and
                  Amanda Askell and
                  Pamela Mishkin and
                  Jack Clark and
                  Gretchen Krueger and
                  Ilya Sutskever},
  editor       = {Marina Meila and
                  Tong Zhang},
  title        = {Learning Transferable Visual Models From Natural Language Supervision},
  booktitle    = {Proceedings of the 38th International Conference on Machine Learning,
                  {ICML} 2021, 18-24 July 2021, Virtual Event},
  series       = {Proceedings of Machine Learning Research},
  volume       = {139},
  pages        = {8748--8763},
  publisher    = {{PMLR}},
  year         = {2021},
  url          = {http://proceedings.mlr.press/v139/radford21a.html},
  bibsource    = {dblp computer science bibliography, https://dblp.org}
}

@inproceedings{DBLP:conf/cvpr/RohrbachRTS15,
  author       = {Anna Rohrbach and
                  Marcus Rohrbach and
                  Niket Tandon and
                  Bernt Schiele},
  title        = {A dataset for Movie Description},
  booktitle    = {{IEEE} Conference on Computer Vision and Pattern Recognition, {CVPR}
                  2015, Boston, MA, USA, June 7-12, 2015},
  pages        = {3202--3212},
  publisher    = {{IEEE} Computer Society},
  year         = {2015},
  url          = {https://doi.org/10.1109/CVPR.2015.7298940},
  doi          = {10.1109/CVPR.2015.7298940},
  bibsource    = {dblp computer science bibliography, https://dblp.org}
}

@inproceedings{DBLP:conf/cvpr/XuMYR16,
  author       = {Jun Xu and
                  Tao Mei and
                  Ting Yao and
                  Yong Rui},
  title        = {{MSR-VTT:} {A} Large Video Description Dataset for Bridging Video
                  and Language},
  booktitle    = {2016 {IEEE} Conference on Computer Vision and Pattern Recognition,
                  {CVPR} 2016, Las Vegas, NV, USA, June 27-30, 2016},
  pages        = {5288--5296},
  publisher    = {{IEEE} Computer Society},
  year         = {2016},
  url          = {https://doi.org/10.1109/CVPR.2016.571},
  doi          = {10.1109/CVPR.2016.571},
  bibsource    = {dblp computer science bibliography, https://dblp.org}
}

@inproceedings{DBLP:conf/iccv/HendricksWSSDR17,
  author       = {Lisa Anne Hendricks and
                  Oliver Wang and
                  Eli Shechtman and
                  Josef Sivic and
                  Trevor Darrell and
                  Bryan C. Russell},
  title        = {Localizing Moments in Video with Natural Language},
  booktitle    = {{IEEE} International Conference on Computer Vision, {ICCV} 2017, Venice,
                  Italy, October 22-29, 2017},
  pages        = {5804--5813},
  publisher    = {{IEEE} Computer Society},
  year         = {2017},
  url          = {https://doi.org/10.1109/ICCV.2017.618},
  doi          = {10.1109/ICCV.2017.618},
  bibsource    = {dblp computer science bibliography, https://dblp.org}
}

@inproceedings{DBLP:conf/iccv/WangWCLWW19,
  author       = {Xin Wang and
                  Jiawei Wu and
                  Junkun Chen and
                  Lei Li and
                  Yuan{-}Fang Wang and
                  William Yang Wang},
  title        = {VaTeX: {A} Large-Scale, High-Quality Multilingual Dataset for Video-and-Language
                  Research},
  booktitle    = {2019 {IEEE/CVF} International Conference on Computer Vision, {ICCV}
                  2019, Seoul, Korea (South), October 27 - November 2, 2019},
  pages        = {4580--4590},
  publisher    = {{IEEE}},
  year         = {2019},
  url          = {https://doi.org/10.1109/ICCV.2019.00468},
  doi          = {10.1109/ICCV.2019.00468},
  bibsource    = {dblp computer science bibliography, https://dblp.org}
}

@inproceedings{DBLP:conf/nips/AlayracRSARFSDZ20,
  author       = {Jean{-}Baptiste Alayrac and
                  Adri{\`{a}} Recasens and
                  Rosalia Schneider and
                  Relja Arandjelovic and
                  Jason Ramapuram and
                  Jeffrey De Fauw and
                  Lucas Smaira and
                  Sander Dieleman and
                  Andrew Zisserman},
  editor       = {Hugo Larochelle and
                  Marc'Aurelio Ranzato and
                  Raia Hadsell and
                  Maria{-}Florina Balcan and
                  Hsuan{-}Tien Lin},
  title        = {Self-Supervised MultiModal Versatile Networks},
  booktitle    = {Advances in Neural Information Processing Systems 33: Annual Conference
                  on Neural Information Processing Systems 2020, NeurIPS 2020, December
                  6-12, 2020, virtual},
  year         = {2020},
  url          = {https://proceedings.neurips.cc/paper/2020/hash/0060ef47b12160b9198302ebdb144dcf-Abstract.html},
  bibsource    = {dblp computer science bibliography, https://dblp.org}
}

@inproceedings{DBLP:conf/icml/RadfordKXBMS23,
  author       = {Alec Radford and
                  Jong Wook Kim and
                  Tao Xu and
                  Greg Brockman and
                  Christine McLeavey and
                  Ilya Sutskever},
  editor       = {Andreas Krause and
                  Emma Brunskill and
                  Kyunghyun Cho and
                  Barbara Engelhardt and
                  Sivan Sabato and
                  Jonathan Scarlett},
  title        = {Robust Speech Recognition via Large-Scale Weak Supervision},
  booktitle    = {International Conference on Machine Learning, {ICML} 2023, 23-29 July
                  2023, Honolulu, Hawaii, {USA}},
  series       = {Proceedings of Machine Learning Research},
  volume       = {202},
  pages        = {28492--28518},
  publisher    = {{PMLR}},
  year         = {2023},
  url          = {https://proceedings.mlr.press/v202/radford23a.html},
  bibsource    = {dblp computer science bibliography, https://dblp.org}
}

@inproceedings{eccv_EclipSE,
  author       = {Yan{-}Bo Lin and
                  Jie Lei and
                  Mohit Bansal and
                  Gedas Bertasius},
  editor       = {Shai Avidan and
                  Gabriel J. Brostow and
                  Moustapha Ciss{\'{e}} and
                  Giovanni Maria Farinella and
                  Tal Hassner},
  title        = {EclipSE: Efficient Long-Range Video Retrieval Using Sight and Sound},
  booktitle    = {Computer Vision - {ECCV} 2022 - 17th European Conference, Tel Aviv,
                  Israel, October 23-27, 2022, Proceedings, Part {XXXIV}},
  series       = {Lecture Notes in Computer Science},
  volume       = {13694},
  pages        = {413--430},
  publisher    = {Springer},
  year         = {2022},
  url          = {https://doi.org/10.1007/978-3-031-19830-4\_24},
  doi          = {10.1007/978-3-031-19830-4\_24},
  bibsource    = {dblp computer science bibliography, https://dblp.org}
}

@inproceedings{audio_enhanced,
  author       = {Sarah Ibrahimi and
                  Xiaohang Sun and
                  Pichao Wang and
                  Amanmeet Garg and
                  Ashutosh Sanan and
                  Mohamed Omar},
  title        = {Audio-Enhanced Text-to-Video Retrieval using Text-Conditioned Feature
                  Alignment},
  booktitle    = {{IEEE/CVF} International Conference on Computer Vision, {ICCV} 2023,
                  Paris, France, October 1-6, 2023},
  pages        = {12020--12030},
  publisher    = {{IEEE}},
  year         = {2023},
  url          = {https://doi.org/10.1109/ICCV51070.2023.01107},
  doi          = {10.1109/ICCV51070.2023.01107},
  bibsource    = {dblp computer science bibliography, https://dblp.org}
}

@article{DBLP:journals/corr/drl-wang,
  author       = {Qiang Wang and
                  Yanhao Zhang and
                  Yun Zheng and
                  Pan Pan and
                  Xian{-}Sheng Hua},
  title        = {Disentangled Representation Learning for Text-Video Retrieval},
  journal      = {CoRR},
  volume       = {abs/2203.07111},
  year         = {2022},
  url          = {https://doi.org/10.48550/arXiv.2203.07111},
  doi          = {10.48550/ARXIV.2203.07111},
  eprinttype    = {arXiv},
  eprint       = {2203.07111},
  bibsource    = {dblp computer science bibliography, https://dblp.org}
}

@inproceedings{DBLP:conf/cvpr/X-PoolGort,
  author       = {Satya Krishna Gorti and
                  No{\"{e}}l Vouitsis and
                  Junwei Ma and
                  Keyvan Golestan and
                  Maksims Volkovs and
                  Animesh Garg and
                  Guangwei Yu},
  title        = {X-Pool: Cross-Modal Language-Video Attention for Text-Video Retrieval},
  booktitle    = {{IEEE/CVF} Conference on Computer Vision and Pattern Recognition,
                  {CVPR} 2022, New Orleans, LA, USA, June 18-24, 2022},
  pages        = {4996--5005},
  publisher    = {{IEEE}},
  year         = {2022},
  url          = {https://doi.org/10.1109/CVPR52688.2022.00495},
  doi          = {10.1109/CVPR52688.2022.00495},
  bibsource    = {dblp computer science bibliography, https://dblp.org}
}

@inproceedings{DBLP:conf/mm/X-CLIPMaXSYZJ22,
  author       = {Yiwei Ma and
                  Guohai Xu and
                  Xiaoshuai Sun and
                  Ming Yan and
                  Ji Zhang and
                  Rongrong Ji},
  editor       = {Jo{\~{a}}o Magalh{\~{a}}es and
                  Alberto Del Bimbo and
                  Shin'ichi Satoh and
                  Nicu Sebe and
                  Xavier Alameda{-}Pineda and
                  Qin Jin and
                  Vincent Oria and
                  Laura Toni},
  title        = {{X-CLIP:} End-to-End Multi-grained Contrastive Learning for Video-Text
                  Retrieval},
  booktitle    = {{MM} '22: The 30th {ACM} International Conference on Multimedia, Lisboa,
                  Portugal, October 10 - 14, 2022},
  pages        = {638--647},
  publisher    = {{ACM}},
  year         = {2022},
  url          = {https://doi.org/10.1145/3503161.3547910},
  doi          = {10.1145/3503161.3547910},
  bibsource    = {dblp computer science bibliography, https://dblp.org}
}

@inproceedings{DBLP:conf/cvpr/bridgeformerGeGLLSQL22,
  author       = {Yuying Ge and
                  Yixiao Ge and
                  Xihui Liu and
                  Dian Li and
                  Ying Shan and
                  Xiaohu Qie and
                  Ping Luo},
  title        = {Bridging Video-text Retrieval with Multiple Choice Questions},
  booktitle    = {{IEEE/CVF} Conference on Computer Vision and Pattern Recognition,
                  {CVPR} 2022, New Orleans, LA, USA, June 18-24, 2022},
  pages        = {16146--16155},
  publisher    = {{IEEE}},
  year         = {2022},
  url          = {https://doi.org/10.1109/CVPR52688.2022.01569},
  doi          = {10.1109/CVPR52688.2022.01569},
  bibsource    = {dblp computer science bibliography, https://dblp.org}
}

@inproceedings{DBLP:conf/iccv/DiffusionRet,
  author       = {Peng Jin and
                  Hao Li and
                  Zesen Cheng and
                  Kehan Li and
                  Xiangyang Ji and
                  Chang Liu and
                  Li Yuan and
                  Jie Chen},
  title        = {DiffusionRet: Generative Text-Video Retrieval with Diffusion Model},
  booktitle    = {{IEEE/CVF} International Conference on Computer Vision, {ICCV} 2023,
                  Paris, France, October 1-6, 2023},
  pages        = {2470--2481},
  publisher    = {{IEEE}},
  year         = {2023},
  url          = {https://doi.org/10.1109/ICCV51070.2023.00234},
  doi          = {10.1109/ICCV51070.2023.00234},
  bibsource    = {dblp computer science bibliography, https://dblp.org}
}

@inproceedings{DBLP:conf/iccv/UATVR,
  author       = {Bo Fang and
                  Wenhao Wu and
                  Chang Liu and
                  Yu Zhou and
                  Yuxin Song and
                  Weiping Wang and
                  Xiangbo Shu and
                  Xiangyang Ji and
                  Jingdong Wang},
  title        = {{UATVR:} Uncertainty-Adaptive Text-Video Retrieval},
  booktitle    = {{IEEE/CVF} International Conference on Computer Vision, {ICCV} 2023,
                  Paris, France, October 1-6, 2023},
  pages        = {13677--13687},
  publisher    = {{IEEE}},
  year         = {2023},
  url          = {https://doi.org/10.1109/ICCV51070.2023.01262},
  doi          = {10.1109/ICCV51070.2023.01262},
  bibsource    = {dblp computer science bibliography, https://dblp.org}
}

@article{DBLP:journals/pami/valor,
  author       = {Jing Liu and
                  Sihan Chen and
                  Xingjian He and
                  Longteng Guo and
                  Xinxin Zhu and
                  Weining Wang and
                  Jinhui Tang},
  title        = {{VALOR:} Vision-Audio-Language Omni-Perception Pretraining Model and
                  Dataset},
  journal      = {{IEEE} Trans. Pattern Anal. Mach. Intell.},
  volume       = {47},
  number       = {2},
  pages        = {708--724},
  year         = {2025},
  url          = {https://doi.org/10.1109/TPAMI.2024.3479776},
  doi          = {10.1109/TPAMI.2024.3479776},
  bibsource    = {dblp computer science bibliography, https://dblp.org}
}

@inproceedings{DBLP:conf/cvpr/querybank,
  author       = {Simion{-}Vlad Bogolin and
                  Ioana Croitoru and
                  Hailin Jin and
                  Yang Liu and
                  Samuel Albanie},
  title        = {Cross Modal Retrieval with Querybank Normalisation},
  booktitle    = {{IEEE/CVF} Conference on Computer Vision and Pattern Recognition,
                  {CVPR} 2022, New Orleans, LA, USA, June 18-24, 2022},
  pages        = {5184--5195},
  publisher    = {{IEEE}},
  year         = {2022},
  url          = {https://doi.org/10.1109/CVPR52688.2022.00513},
  doi          = {10.1109/CVPR52688.2022.00513},
  bibsource    = {dblp computer science bibliography, https://dblp.org}
}

@article{DBLP:journals/corr/dual_softmax,
  author       = {Xing Cheng and
                  Hezheng Lin and
                  Xiangyu Wu and
                  Fan Yang and
                  Dong Shen},
  title        = {Improving Video-Text Retrieval by Multi-Stream Corpus Alignment and
                  Dual Softmax Loss},
  journal      = {CoRR},
  volume       = {abs/2109.04290},
  year         = {2021},
  url          = {https://arxiv.org/abs/2109.04290},
  eprinttype    = {arXiv},
  eprint       = {2109.04290},
  bibsource    = {dblp computer science bibliography, https://dblp.org}
}

@inproceedings{DBLP:conf/nips/AkbariYQCCCG21,
  author       = {Hassan Akbari and
                  Liangzhe Yuan and
                  Rui Qian and
                  Wei{-}Hong Chuang and
                  Shih{-}Fu Chang and
                  Yin Cui and
                  Boqing Gong},
  editor       = {Marc'Aurelio Ranzato and
                  Alina Beygelzimer and
                  Yann N. Dauphin and
                  Percy Liang and
                  Jennifer Wortman Vaughan},
  title        = {{VATT:} Transformers for Multimodal Self-Supervised Learning from
                  Raw Video, Audio and Text},
  booktitle    = {Advances in Neural Information Processing Systems 34: Annual Conference
                  on Neural Information Processing Systems 2021, NeurIPS 2021, December
                  6-14, 2021, virtual},
  pages        = {24206--24221},
  year         = {2021},
  url          = {https://proceedings.neurips.cc/paper/2021/hash/cb3213ada48302953cb0f166464ab356-Abstract.html},
  bibsource    = {dblp computer science bibliography, https://dblp.org}
}

@article{DBLP:journals/corr/abs-1804-02516,
  author       = {Antoine Miech and
                  Ivan Laptev and
                  Josef Sivic},
  title        = {Learning a Text-Video Embedding from Incomplete and Heterogeneous
                  Data},
  journal      = {CoRR},
  volume       = {abs/1804.02516},
  year         = {2018},
  url          = {http://arxiv.org/abs/1804.02516},
  eprinttype    = {arXiv},
  eprint       = {1804.02516},
  bibsource    = {dblp computer science bibliography, https://dblp.org}
}

@article{hunyuan,
  title={Tencent text-video retrieval: hierarchical cross-modal interactions with multi-level representations},
  author={Jiang, Jie and Min, Shaobo and Kong, Weijie and Wang, Hongfa and Li, Zhifeng and Liu, Wei},
  journal={IEEE Access},
  year={2022},
  publisher={IEEE}
}

@inproceedings{DBLP:conf/sigir/centerclip,
  author       = {Shuai Zhao and
                  Linchao Zhu and
                  Xiaohan Wang and
                  Yi Yang},
  editor       = {Enrique Amig{\'{o}} and
                  Pablo Castells and
                  Julio Gonzalo and
                  Ben Carterette and
                  J. Shane Culpepper and
                  Gabriella Kazai},
  title        = {CenterCLIP: Token Clustering for Efficient Text-Video Retrieval},
  booktitle    = {{SIGIR} '22: The 45th International {ACM} {SIGIR} Conference on Research
                  and Development in Information Retrieval, Madrid, Spain, July 11 -
                  15, 2022},
  pages        = {970--981},
  publisher    = {{ACM}},
  year         = {2022},
  url          = {https://doi.org/10.1145/3477495.3531950},
  doi          = {10.1145/3477495.3531950},
  bibsource    = {dblp computer science bibliography, https://dblp.org}
}

@inproceedings{DBLP:conf/iclr/InternVid,
  author       = {Yi Wang and
                  Yinan He and
                  Yizhuo Li and
                  Kunchang Li and
                  Jiashuo Yu and
                  Xin Ma and
                  Xinhao Li and
                  Guo Chen and
                  Xinyuan Chen and
                  Yaohui Wang and
                  Ping Luo and
                  Ziwei Liu and
                  Yali Wang and
                  Limin Wang and
                  Yu Qiao},
  title        = {InternVid: {A} Large-scale Video-Text Dataset for Multimodal Understanding
                  and Generation},
  booktitle    = {The Twelfth International Conference on Learning Representations,
                  {ICLR} 2024, Vienna, Austria, May 7-11, 2024},
  publisher    = {OpenReview.net},
  year         = {2024},
  url          = {https://openreview.net/forum?id=MLBdiWu4Fw},
  bibsource    = {dblp computer science bibliography, https://dblp.org}
}

@inproceedings{DBLP:conf/nips/VAST,
  author       = {Sihan Chen and
                  Handong Li and
                  Qunbo Wang and
                  Zijia Zhao and
                  Mingzhen Sun and
                  Xinxin Zhu and
                  Jing Liu},
  editor       = {Alice Oh and
                  Tristan Naumann and
                  Amir Globerson and
                  Kate Saenko and
                  Moritz Hardt and
                  Sergey Levine},
  title        = {{VAST:} {A} Vision-Audio-Subtitle-Text Omni-Modality Foundation Model
                  and Dataset},
  booktitle    = {Advances in Neural Information Processing Systems 36: Annual Conference
                  on Neural Information Processing Systems 2023, NeurIPS 2023, New Orleans,
                  LA, USA, December 10 - 16, 2023},
  year         = {2023},
  url          = {http://papers.nips.cc/paper\_files/paper/2023/hash/e6b2b48b5ed90d07c305932729927781-Abstract-Conference.html},
  bibsource    = {dblp computer science bibliography, https://dblp.org}
}

@inproceedings{DBLP:conf/cvpr/JeongPKK25,
  author       = {Boseung Jeong and
                  Jicheol Park and
                  Sungyeon Kim and
                  Suha Kwak},
  title        = {Learning Audio-guided Video Representation with Gated Attention for
                  Video-Text Retrieval},
  booktitle    = {{IEEE/CVF} Conference on Computer Vision and Pattern Recognition,
                  {CVPR} 2025, Nashville, TN, USA, June 11-15, 2025},
  pages        = {26202--26211},
  publisher    = {Computer Vision Foundation / {IEEE}},
  year         = {2025},
  url          = {https://openaccess.thecvf.com/content/CVPR2025/html/Jeong\_Learning\_Audio-guided\_Video\_Representation\_with\_Gated\_Attention\_for\_Video-Text\_Retrieval\_CVPR\_2025\_paper.html},
  bibsource    = {dblp computer science bibliography, https://dblp.org}
}

@inproceedings{DBLP:conf/nips/BaevskiZMA20,
  author       = {Alexei Baevski and
                  Yuhao Zhou and
                  Abdelrahman Mohamed and
                  Michael Auli},
  editor       = {Hugo Larochelle and
                  Marc'Aurelio Ranzato and
                  Raia Hadsell and
                  Maria{-}Florina Balcan and
                  Hsuan{-}Tien Lin},
  title        = {wav2vec 2.0: {A} Framework for Self-Supervised Learning of Speech
                  Representations},
  booktitle    = {Advances in Neural Information Processing Systems 33: Annual Conference
                  on Neural Information Processing Systems 2020, NeurIPS 2020, December
                  6-12, 2020, virtual},
  year         = {2020},
  url          = {https://proceedings.neurips.cc/paper/2020/hash/92d1e1eb1cd6f9fba3227870bb6d7f07-Abstract.html},
  bibsource    = {dblp computer science bibliography, https://dblp.org}
}

@article{DBLP:journals/corr/excae,
  author       = {Junxiang Chen and
                  Baoyao Yang and
                  Wenbin Yao},
  title        = {Expertized Caption Auto-Enhancement for Video-Text Retrieval},
  journal      = {CoRR},
  volume       = {abs/2502.02885},
  year         = {2025},
  url          = {https://doi.org/10.48550/arXiv.2502.02885},
  doi          = {10.48550/ARXIV.2502.02885},
  eprinttype    = {arXiv},
  eprint       = {2502.02885},
  bibsource    = {dblp computer science bibliography, https://dblp.org}
}

@inproceedings{gemmeke2017audioset,
  title={AudioSet: An ontology and human-labeled dataset for audio events},
  author={Gemmeke, J F and Ellis, D P and Freedman, D and Jansen, A and Lawrence, W and Moore, C and Plakal, M and Ritter, M},
  booktitle={2017 IEEE International Conference on Acoustics, Speech and Signal Processing (ICASSP)},
  pages={776--780},
  year={2017},
  organization={IEEE}
}

\clearpage
\setcounter{page}{1}
\maketitlesupplementary


\section{Appendix}
This supplementary material provides additional details of GAIS's architectural and experimental results, which we could not include in the main paper.
\subsection{More Architectural Details}\label{sec:arch_details}
To enable frame-level audio–visual feature fusion, we first align audio representations with the dimensionality of video frame features. For the Whisper model \cite{DBLP:conf/icml/RadfordKXBMS23}, raw audio is sampled at 16kHz, converted to waveform features, and subsequently transformed into log-Mel spectrograms as input to the model. Since Whisper supports audio sequences of up to 30 seconds, longer clips are concatenated and resampled to the target temporal resolution. After processing, the base model produces features of size [1500 × 512], while the small model outputs [1500 × 768].

For Wav2Vec2.0, which supports arbitrary-length audio input, we adopt a similar preprocessing pipeline. The base version generates features of [1500 × 768], and the large version produces [1500 × 1024]. These representations are subsequently aligned with video frame features for downstream fusion.

\subsection{More Implementation Details}\label{sec:impl_details}
The details of the training configurations of our method across datasets are provided in \cref{tab:configuration}. We follow T-MASS \cite{wang_text_2024} for most configurations, such as the image encoder, optimizer, Transformer dropout, support loss weight and Learning rate for Non-CLIP parameters.
\begin{table}[htbp]
\fontsize{9}{10}\selectfont
    \centering
    \setlength{\tabcolsep}{1mm}
    \begin{tabular}{l|cccc}
    \toprule
    Source Dataset    &  MSR-VTT & DiDemo & LSMDC & VATEX \\
    \midrule
    Image encoder & \multicolumn{4}{c}{CLIP-ViTs (B/32 and B/16)} \\
    Audio encoder & \multicolumn{4}{c}{Whisper [base]} \\
    Total epochs  & \multicolumn{4}{c}{5} \\
    Optimizer & \multicolumn{4}{c}{Adam} \\
    Batch size  & \multicolumn{4}{c}{32} \\
    Max frames & \multicolumn{4}{c}{12} \\
    Transformer dropout & 0.3 & 0.4 & 0.3 & 0.4 \\
    Support loss weight & 0.8 & 0.1 & 0.3 & 0.4 \\
    
    \bottomrule
    \end{tabular}
    \caption{Training configurations of various datasets}
    \label{tab:configuration}
\end{table}

\subsection{Audio Bucket Statistics}\label{sec:audio_bucket}

\begin{figure}[t]
    \centering
    \includegraphics[width=0.9\linewidth]{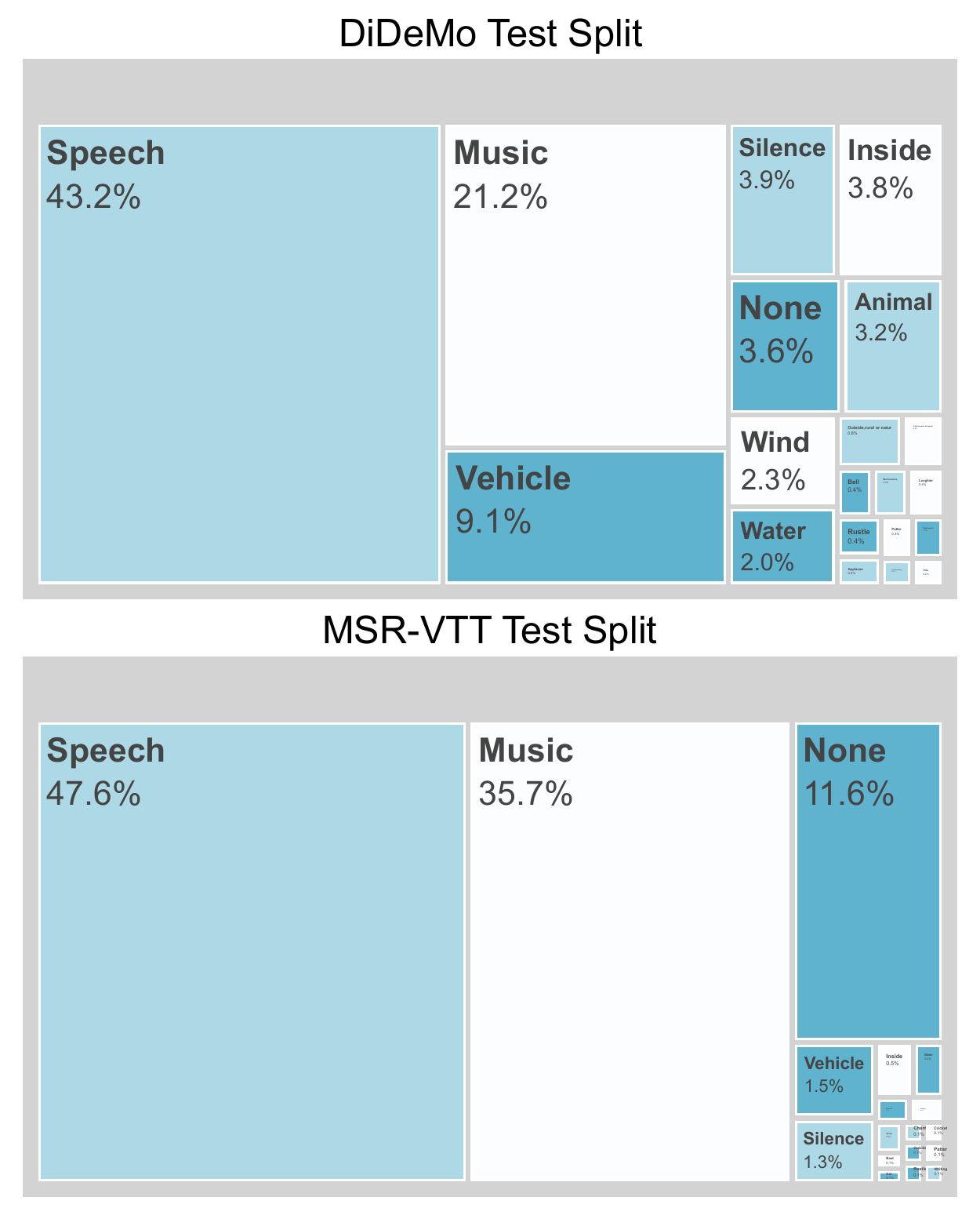}
    \caption{Audio type treemap of different datasets.}
    \label{fig:audio_treemap}
\end{figure}

To better understand how audio contributes to text–video retrieval, we categorize videos by their dominant audio type and report the distribution of audio categories in MSR-VTT and DiDeMo. Only categories with at least 20 samples are included in our main bucket-level evaluation. As shown in the treemap visualizations (\cref{fig:audio_treemap}), Speech and Music are the two most frequent audio types in both datasets, jointly accounting for over 60$\%$ of all videos. The remaining categories form a long-tail distribution, with many containing fewer than 10 samples. This imbalance explains why improvements are most evident in the high-frequency buckets: GAIS leverages audio selectively, and semantic audio cues are more reliable when sufficient training instances exist.

\subsection{More Quantitative Results}\label{sec:more_quanti_result}

\textbf{Effect of Post process.} 
In text–video retrieval tasks, post-processing techniques have been widely adopted to enhance retrieval performance. Many prior methods leverage strategies such as Dual Softmax Loss (DSL) and Querybank Normalization (QB-Norm) to achieve significant improvements. In our experiments, we apply DSL as a post-processing step during inference. Notably, on the MSR-VTT dataset, incorporating DSL leads to an R@1 improvement of up to 6.9\%, demonstrating its remarkable effectiveness.

We observe that Dual Softmax Loss (DSL) yields the most noticeable improvements when the raw similarity scores are unevenly distributed, in large-scale retrieval scenarios, or on datasets with highly similar video content (e.g., MSR-VTT). In these cases, DSL’s bidirectional normalization better highlights high-confidence matches and suppresses noisy candidates.

\begin{table}[h]
\fontsize{9}{10}\selectfont
    \centering
    \setlength{\tabcolsep}{1.5mm}
    \begin{tabular}{l|l l l l l}
        \toprule
        Method & R@1$\uparrow$ & R@5$\uparrow$ & R@10$\uparrow$ & MnR$\downarrow$ \\
         \midrule
         CAMoE\cite{DBLP:journals/corr/dual_softmax} & 44.6 & 72.6 & 81.8  & 13.3 \\
         \qquad+DSL & 47.3 \textcolor{blue}{(+2.7)} & 74.2 \textcolor{blue}{(+1.6)} & 84.5 \textcolor{blue}{(+2.7)} & 11.9 \\
         TS2-Net\cite{liu_ts2-net_2022} &47.0   &74.5    &83.8       &13.0 \\
         \qquad+DSL & 51.1 \textcolor{blue}{(+4.1)} & 76.9 \textcolor{blue}{(+2.4)} & 85.6 \textcolor{blue}{(+1.8)}  & 9.2 \\
         UATVR\cite{DBLP:conf/iccv/UATVR} &47.5   &73.9    &83.5      &12.3 \\
         \qquad+DSL & 49.8 \textcolor{blue}{(+2.3)}   &76.1 \textcolor{blue}{(+2.2)}   &85.5 \textcolor{blue}{(+2.0)}       &12.9    \\
         TEFAL\cite{audio_enhanced} &49.9   &76.2    &83.5      &11.4 \\
         \qquad+DSL & 50.1 \textcolor{blue}{(+0.2)}   &77.0 \textcolor{blue}{(+0.8)}   &85.5 \textcolor{blue}{(+2.0)}       &10.5    \\
         AVIGATE\cite{DBLP:conf/cvpr/JeongPKK25} & 50.2   &74.3   &83.2     &- \\
         \qquad+DSL &53.9 \textcolor{blue}{(+3.7)} & 77.0 \textcolor{blue}{(+2.7)} &86.0 \textcolor{blue}{(+2.8)}  &- \\
         \rowcolor{gray!20}
         GAIS(Ours) & 57.0 &83.1   &90.9     &7.6 \\
         \rowcolor{gray!20}
         \qquad+DSL & \textbf{63.9} \textcolor{blue}{(+6.9)} &\textbf{86.7} \textcolor{blue}{(+3.6)} &\textbf{93.1} \textcolor{blue}{(+2.2)}  &\textbf{6.2} \\
        \bottomrule
    \end{tabular}
    
    \caption{Text-to-Video retrieval results on the MSR-VTT 9k split. The post-processing techniques such as DSL or QB-Norm are used for further performance boosting.}
    \label{tab:dsl_res}
\end{table}

\begin{figure}[h]
    \centering
    \includegraphics[width=1\linewidth]{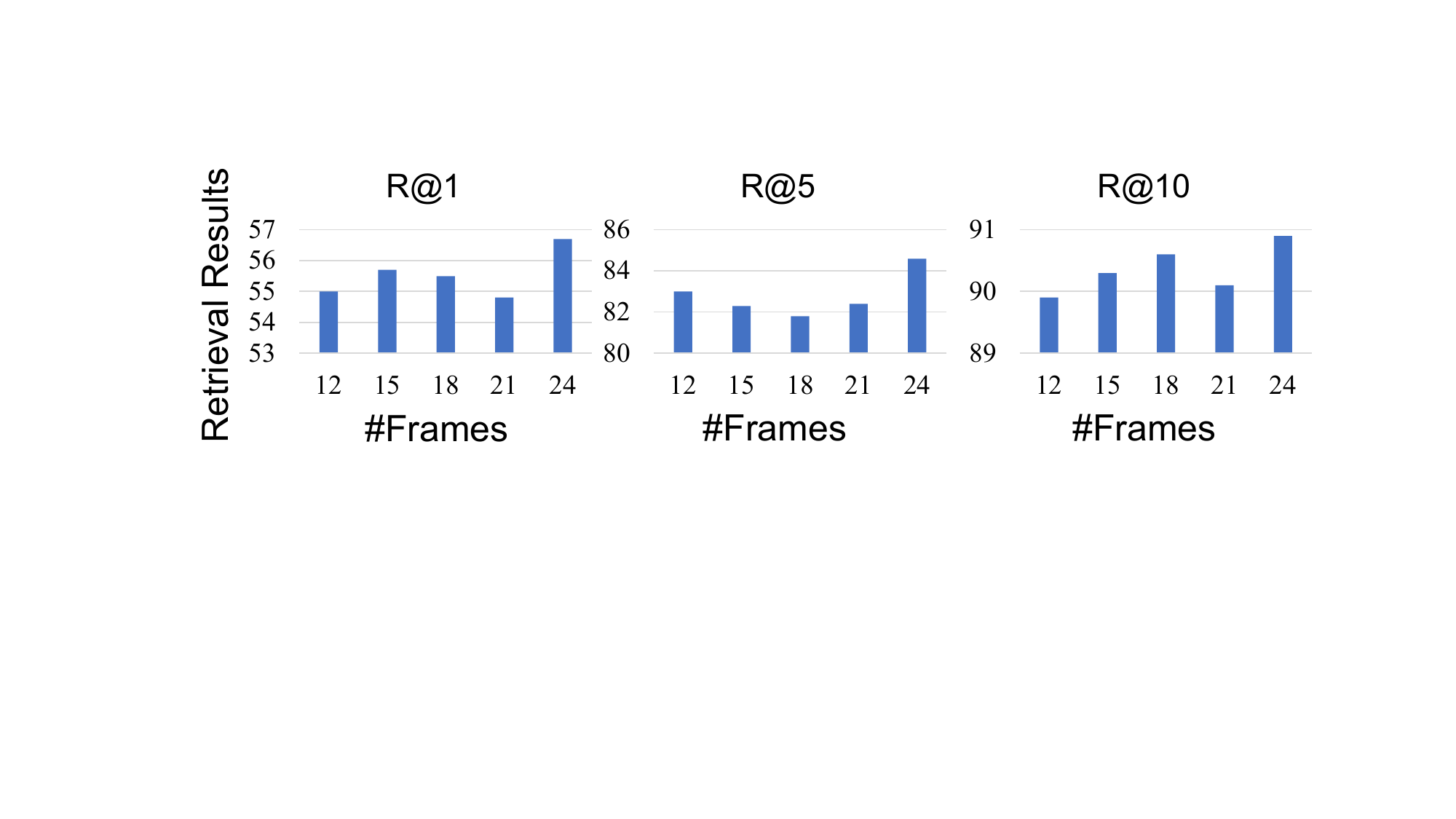}
    \caption{Effect of sampled frame number on text-to-video retrieval performance (MSR-VTT 9k split). GAIS consistently improves with denser frame sampling due to frame-level audio-visual gating.}
    \label{fig:frames_compare}
\end{figure}

\textbf{Effect of Frame Number.} We also discuss the effect of the \#frames in Fig. \ref{fig:frames_compare}. Specifically, we report performance with frames = \{12, 15, 18, 21, 24\}. GAIS enables a notable performance boost with denser frame sampling. Benefiting from the frame-level gating mechanism in our audio–visual fusion, our method exhibits consistent performance gains with more sampled frames. For fair comparison with previous approaches, we fix the number of sampled frames to 12 per video for all datasets.

\textbf{Effect of Supprt loss weight.} To evaluate the influence of the support-based refinement term, we vary the loss weight $\lambda\in[0,1]$ and examine its effect on retrieval accuracy. \cref{tab:support_list_weight} reports the effect of varying the support loss weight $\lambda$ on the MSR-VTT 9k split. We observe a clear rise–then–fall trend across R@1/5/10 as $\lambda$ increases from 0 to 1. Small to moderate values $\lambda \leq 0.8$ consistently improve retrieval accuracy, indicating that the support-based refinement provides useful margin stabilization without overpowering the main contrastive objective. However, further increasing $\lambda$ to 1.0 leads to performance degradation, suggesting that excessively strong refinement may restrict representation flexibility. Based on these results, we use $\lambda=0.8$ as the default setting for MSR-VTT.

\begin{table}[h]
    \centering
    \begin{tabular}{l|cccc>{\columncolor{gray!20}}cc}
        \toprule
        $\lambda$ & 0 & 0.2 & 0.4 & 0.6 & 0.8 & 1.0 \\  
        \midrule
        R@1 & 54.1 & 57.1 & 57.0 & 56.6  & 57.0 & 55.7 \\
        R@5 & 80.9 & 82.8 & 82.6 & 83.1 & 83.1 & 83.1 \\
        R@10 & 89.6 & 89.9 & 90.5 & 90.7  & 90.9 & 90.4 \\
        MnR & 7.71 & 7.94 & 7.78 & 7.65 & 7.63 & 7.59 \\
        \bottomrule
    \end{tabular}
    \caption{Effect of Support loss weight $\lambda$ on MSR-VTT 9k split.}
    \label{tab:support_list_weight}
\end{table}

\begin{figure}[ht]
    \centering
    \includegraphics[width=1\linewidth]{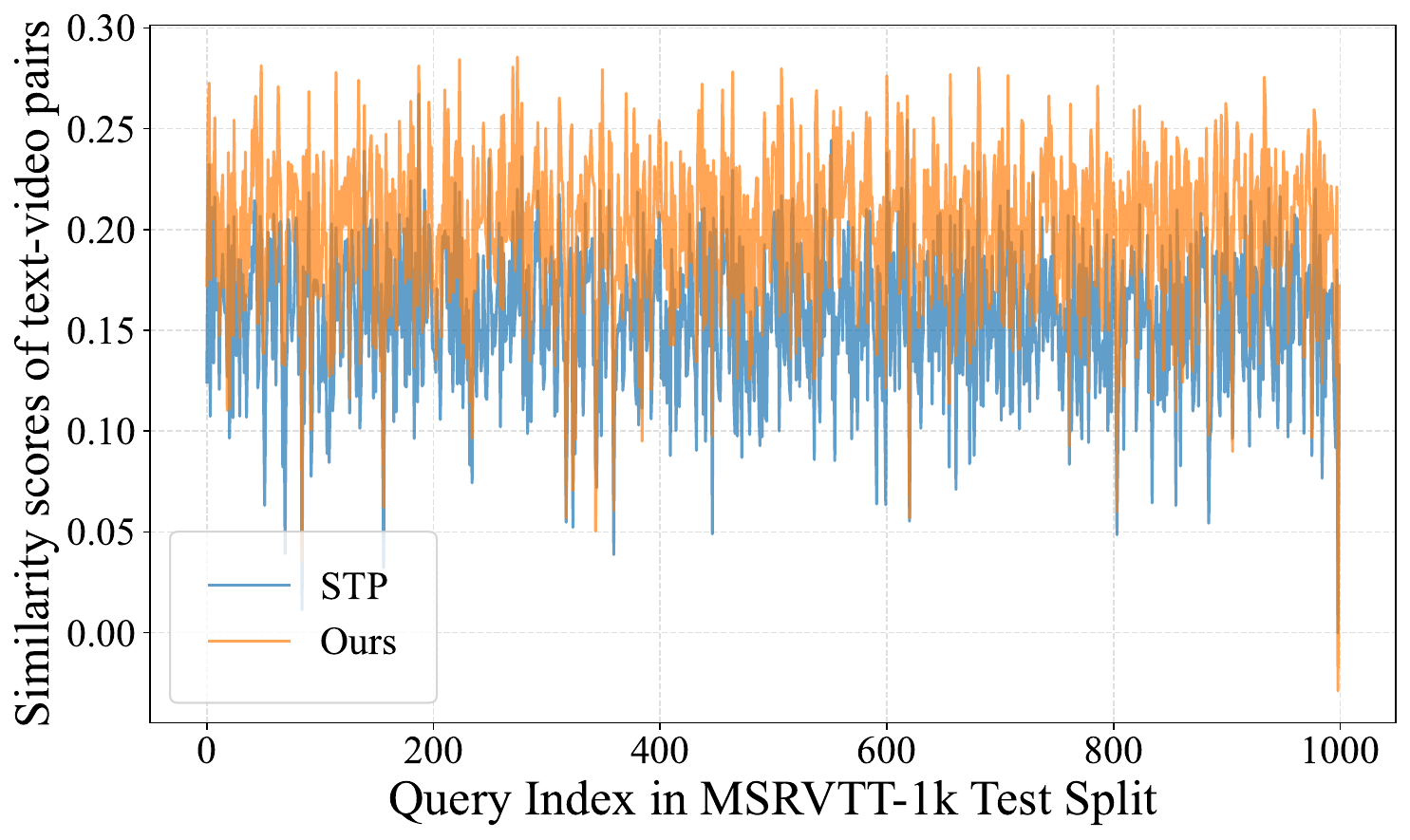}
    \caption{Comparison of cosine similarity distributions between naive stochastic text perturbation (STP) and SVSP on MSR-VTT 1k test split. SVSP produces higher and more stable similarity scores for relevant text-video pairs, demonstrating improved robustness and alignment.}
    \label{fig:SVSP_similarity}
\end{figure}

\textbf{The similarity under STP and SVSP.} To evaluate the robustness of perturbation strategies, we compare the cosine similarity scores between encoded texts and their ground-truth videos under stochastic text perturbation (STP) and SVSP on the MSRVTT 1k test split. As shown in \cref{fig:SVSP_similarity}, STP produces large fluctuations in similarity values due to its isotropic and uncontrolled noise injection, resulting in unstable retrieval behavior. In contrast, SVSP yields consistently higher and smoother similarity curves, indicating that variance-scaled and directionally aligned perturbations produce a more coherent embedding neighborhood. This empirical evidence complements the main ablation results and further demonstrates the robustness advantage of SVSP.

\subsection{More Qualitative Results}\label{sec:more_quali_result}

\begin{figure*}[t]
    \centering
    \includegraphics[width=1\linewidth]{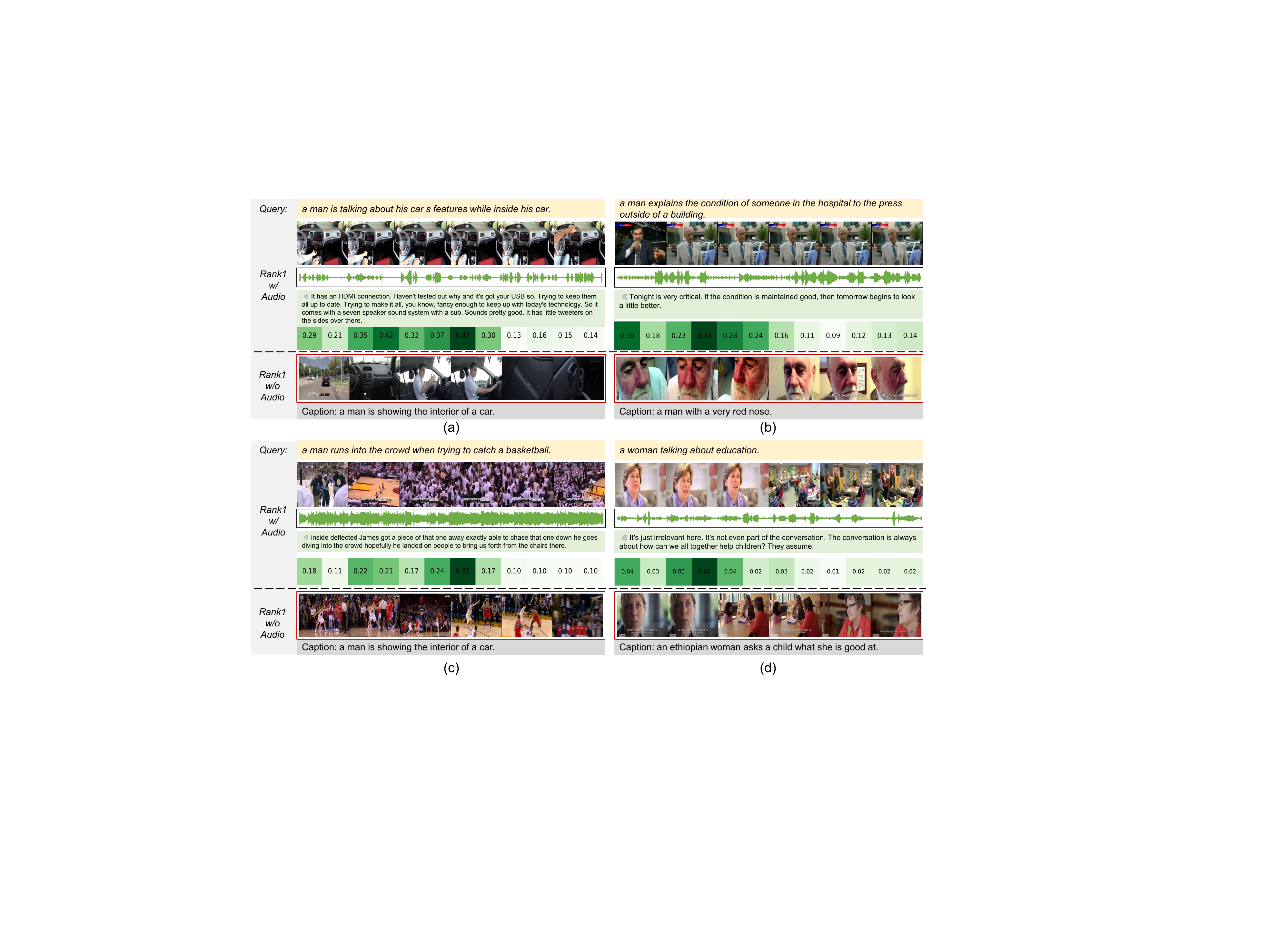}
    \caption{Additional qualitative results of text-to-video retrieval. Each example shows the video caption, the text query, and the retrieved audio transcript under our method and its audio-ablated variant (Ours w/o Audio).
These results demonstrate that incorporating audio cues improves semantic alignment, particularly for queries involving speech or context-sensitive sounds. }
    \label{fig:appendix_results}
\end{figure*}

We provide additional qualitative examples in \cref{fig:appendix_results} to further illustrate how GAIS leverages audio information for text-to-video retrieval. These examples compare the retrieved top-1 results with and without audio. When speech or context-relevant acoustic cues are present, GAIS successfully attends to frames aligned with the transcript and retrieves semantically accurate videos, while the audio-ablated variant often falls back to visually similar but semantically mismatched candidates. The gating weights highlight the model’s ability to emphasize informative dialogue and suppress irrelevant background noise. These results demonstrate that audio cues substantially enhance semantic alignment, particularly for queries involving speech, conversations, or context-sensitive sounds.

\begin{figure*}[h]
    \centering
    \includegraphics[width=1\linewidth]{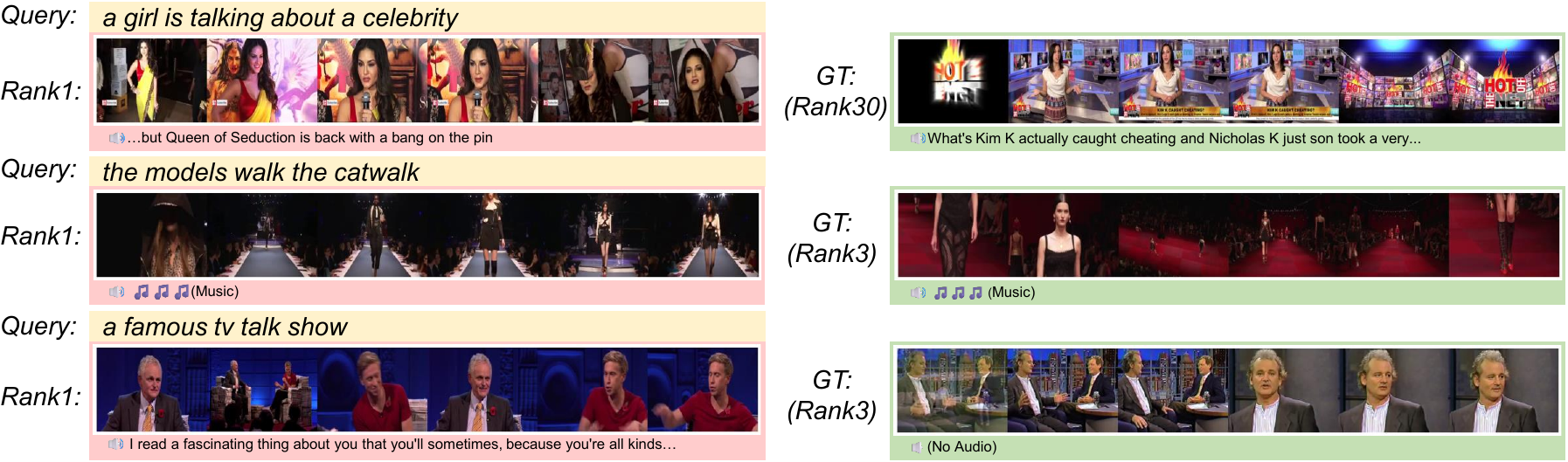}
    \caption{Typical failure cases categorized into visually similar scenes, semantically similar audio cues, and missing-audio ground truths.}
    \label{fig:failure_cases}
\end{figure*}

\subsubsection{Failure Case Analysis}\label{sec:failure_case}
To better understand the limitations of our model, we further analyze representative failure cases, which can be roughly divided into three categories, as illustrated in \cref{fig:failure_cases}.
(1) Visually similar scenes with identical audio context: when multiple clips share highly similar visual elements and background music, the model struggles to distinguish them.
(2) Semantically similar audio cues: when the audio content conveys similar semantics (e.g., people talking, ambient chatter), the model’s retrieval confidence may blur across candidates.
(3) Missing-audio ground truth: in cases where the ground-truth video lacks audio, our model tends to favor visually similar clips that include auxiliary audio cues.
These cases highlight the challenges of fine-grained audio-visual alignment under ambiguous or missing audio conditions.

\end{document}